\documentclass[lettersize,journal]{IEEEtran}
\usepackage{amsmath,amsfonts}
\usepackage{algorithmic}
\usepackage{algorithm}
\usepackage{array}
\usepackage[caption=false,font=normalsize,labelfont=sf,textfont=sf]{subfig}
\usepackage{textcomp}
\usepackage{stfloats}
\usepackage{url}
\usepackage{indentfirst}
\usepackage{xcolor}
\usepackage{bm}
\usepackage{verbatim}

\usepackage{graphicx}
\usepackage{cite}

\usepackage{color}
\newcommand{\revise}[1]{\textcolor{black}{#1}}
\usepackage{multirow}
\usepackage{multicol}
\usepackage{array}
\usepackage{longtable}
\usepackage{bbding} 

\usepackage{booktabs}
\usepackage{colortbl}  
\usepackage{xcolor}
\usepackage{lipsum}

\usepackage[colorlinks,
            linkcolor=red,
            anchorcolor=blue,
            citecolor=green]{hyperref}
\usepackage{ulem}
\begin{document}
\title{RoSe: Robust Self-supervised Stereo Matching under Adverse Weather Conditions}

\author{Yun Wang, Junjie Hu, Junhui Hou, Chenghao Zhang, Renwei Yang, Dapeng Oliver Wu*
       
\thanks{
This work was supported by the InnoHK Initiative of the Government of the Hong Kong SAR and the Laboratory for Artificial Intelligence (AI)-Powered Financial Technologies, with additional support from the Hong Kong Research Grants Council (RGC) grant C1042-23GF and the Hong Kong Innovation and Technology Fund (ITF) grant MHP/061/23. (\textit{Corresponding author: Dapeng Oliver Wu.})

Yun Wang and Renwei Yang are with the Department of Computer Science, City University of Hong Kong, Kowloon, Hong Kong SAR, China (e-mail: ywang3875-c@my.cityu.edu.hk, renwei.y@my.cityu.edu.hk). 

Junjie Hu is with the Chinese University of Hong Kong, Shenzhen, China. (Email: hujunjie@cuhk.edu.cn).

Chenghao Zhang is with the Institute of Automation, Chinese Academy of Sciences (CASIA). (Email: zhangchenghao18@mails.ucas.ac.cn)

Junhui Hou and Dapeng Oliver Wu are with the Department of Computer Science, City University of Hong Kong, Kowloon, Hong Kong SAR, China (e-mail: jh.hou@cityu.edu.hk, dpwu@ieee.org).
}}

\markboth{Journal of \LaTeX\ Class Files,~Vol.~14, No.~8, September~2023}%
{Shell \MakeLowercase{\textit{et al.}}: A Sample Article Using IEEEtran.cls for IEEE Journals}

\maketitle

\begin{abstract}
Recent self-supervised stereo matching methods have made significant progress, but their performance significantly degrades under adverse weather conditions such as night, rain, and fog. We identify two primary weaknesses contributing to this performance degradation. First, adverse weather introduces noise and reduces visibility, making CNN-based feature extractors struggle with degraded regions like reflective and textureless areas. Second, these degraded regions can disrupt accurate pixel correspondences, leading to ineffective supervision based on the photometric consistency assumption.
To address these challenges, we propose injecting robust priors derived from the visual foundation model into the CNN-based feature extractor to improve feature representation under adverse weather conditions. We then introduce scene correspondence priors to construct robust supervisory signals rather than relying solely on the photometric consistency assumption.
Specifically, we create synthetic stereo datasets with realistic weather degradations. These datasets feature clear and adverse image pairs that maintain the same semantic context and disparity, preserving the scene correspondence property.
With this knowledge, we propose a robust self-supervised training paradigm, consisting of two key steps: robust self-supervised scene correspondence learning and adverse weather distillation. Both steps aim to align underlying scene results from clean and adverse image pairs, thus improving model disparity estimation under adverse weather effects.
Extensive experiments demonstrate the effectiveness and versatility of our proposed solution, which outperforms existing state-of-the-art self-supervised methods.
Codes are available at \textcolor{blue}{https://github.com/cocowy1/RoSe-Robust-Self-supervised-Stereo-Matching-under-Adverse-Weather-Conditions}.
\end{abstract}

\begin{IEEEkeywords}
Stereo Matching, Self-supervised Learning, Depth Estimation, Adverse Weather Conditions
\end{IEEEkeywords}

\section{Introduction}
\label{sec1:intro}
Disparity estimation from stereo images is a critical task in autonomous driving and scene reconstruction. While deep supervised stereo matching approaches have yielded impressive results, they rely on expensive ground truth data and are limited to small-scale real-world datasets. In contrast, self-supervised methods overcome these constraints by eliminating the need for ground truth data and offer the potential to scale data usage to billions of samples.

\begin{figure}[t]
    \includegraphics[width=1.0\linewidth]{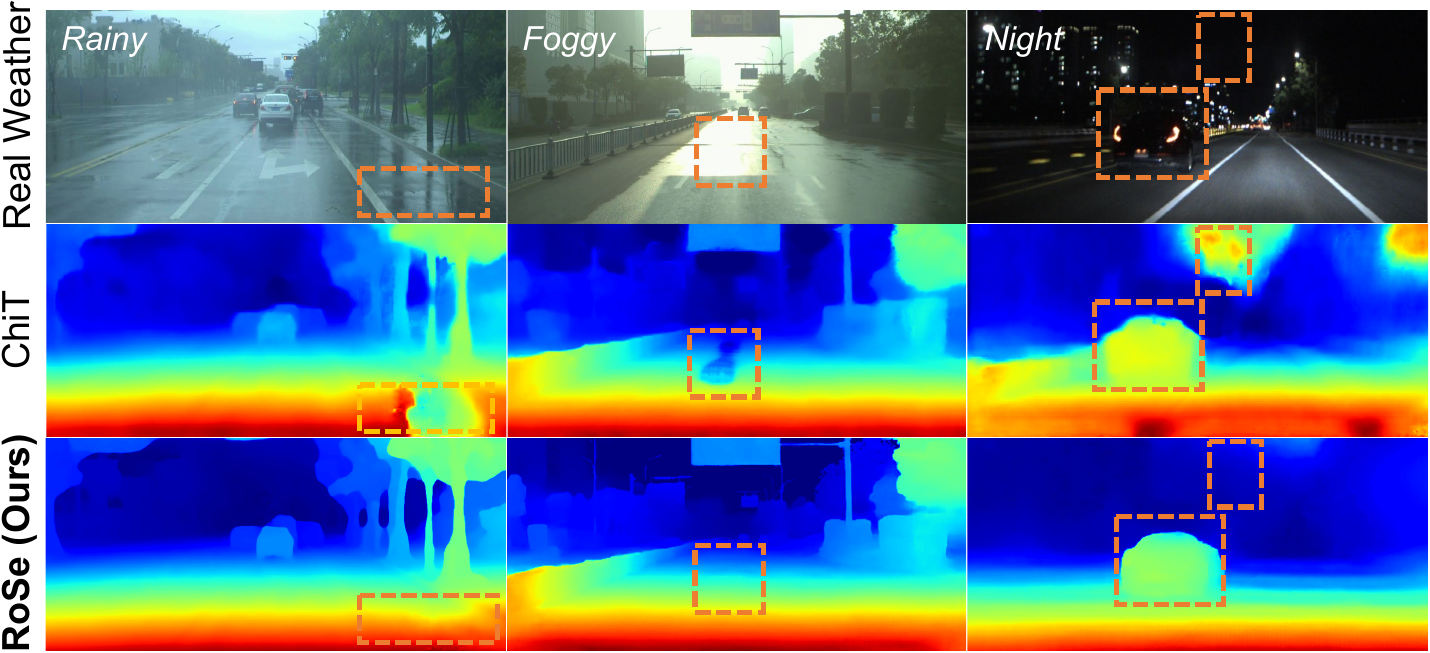}
    \vspace{-0.6cm}
    \caption{Visual comparison with self-supervised SoTA method ChiT~\cite{su2022chitransformer} under challenging conditions. Standard self-supervised approaches struggle with \textit{reflective, blur, and textureless regions} due to the photometric consistency assumption. 
    In contrast, our model reliably estimates in adverse conditions.}
    \label{sec1:teaser}
    \vspace{-0.3cm}
\end{figure}

\begin{figure*}[t]
\centering
\includegraphics[width=1.\textwidth]{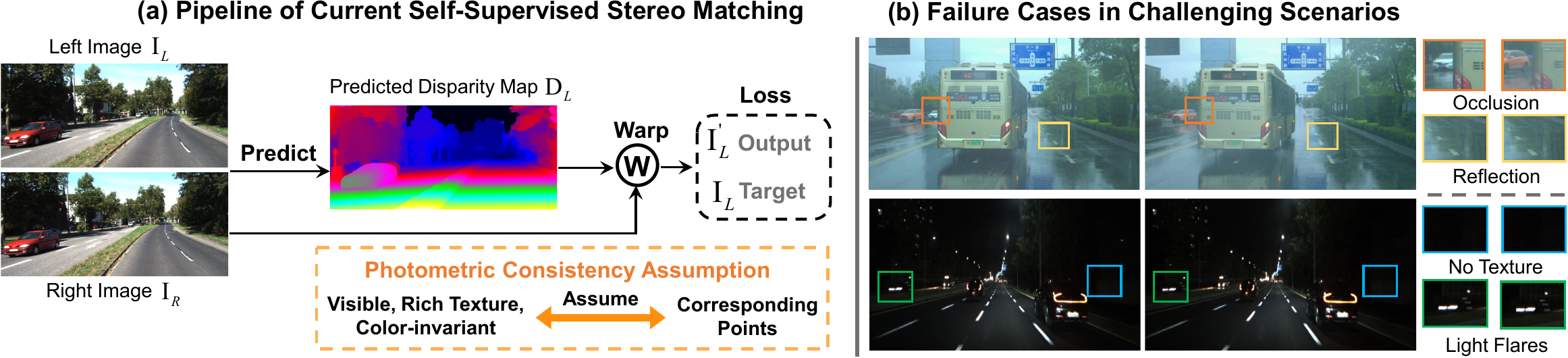}
\vspace{-0.6cm}
\caption{(a) Illustration of existing self-supervised stereo matching methods training pipeline.
The photometric consistency assumption is to maximize the similarity between the left image ${\bf{I}}_{L}$ and the reconstructed left image ${\bf I}^{'}_{L}$ after being warped from the right image ${\bf I}_{R}$.
(b) However, this assumption is weak in adverse weather scenarios and many real-world disturbances would violate the photometric consistency assumption, leading to ambiguous supervision.}
\vspace{-0.3cm}
\label{sec1:assumption}
\end{figure*}

However, self-supervised methods experience significant performance degradation under adverse lighting and weather conditions, such as night, rain, and fog, as illustrated in Fig.~\ref{sec1:teaser}. 
We identify two key weaknesses that hinder their performance under these adverse conditions:
(i) the limited capacity of CNN-based feature extractors.
Existing self-supervised stereo matching methods typically utilize CNN-based Feature Pyramid Networks (FPNs)~\cite{wang2020parallax, liu2020flow2stereo, zhang2019dispsegnet, su2022chitransformer, zhang2023unsupervised,wang2025dualnet} for feature extraction. 
However, CNNs with limited receptive fields struggle to extract discriminative features in challenging regions such as reflections and textureless areas~\cite{ramirez2022open}.
(ii) ineffective supervision from the photometric consistency assumption.
Adverse weather conditions introduce noise and reduce visibility, which can disrupt accurate pixel correspondences across views, potentially leading to ineffective supervision based on the photometric consistency assumption, as shown in Fig.~\ref{sec1:assumption}.


\revise{To enhance feature distinctiveness under challenging conditions, we turn to recent Vision Foundation Models. In particular, we choose SAM~\cite{kirillov2023segment} and DAMv2~\cite{depth_anything_v2} due to their robustness in dense prediction tasks, which are better aligned with the demands of stereo matching than image- or object-level recognition tasks.}
However, simply incorporating pre-trained VFMs into stereo matching may not yield optimal results. Unlike Feature Pyramid Networks (FPNs), VFMs typically employ a plain Vision Transformer (ViT) architecture~\cite{dosovitskiy2020image}, resulting in features extracted at a lower resolution (e.g., 1/16), which may lead to a loss of image details. Moreover, these VFMs, trained for single-image tasks, excel at extracting high-level semantic features but lack sufficient stereo correspondence measurements in dense cross-view feature matching~\cite{weinzaepfel2023croco}.
Therefore, we propose integrating robust priors from VFMs into the FPN to combine the strengths of CNNs and ViTs. Additionally, several studies~\cite{qiao2023robustness,wang2024empirical,chen2024robustsam} have shown that VFMs like SAM degrade under adverse conditions such as low lighting, and adverse weather, affecting their real-world robustness. 
\revise{To address this, we propose an Anti-Adverse Feature Enhancement Module (AFEM) designed to suppress image style attributes introduced by adverse weather while preserving core content information. AFEM operates across the spatial, channel, and frequency domains, effectively disentangling degradation-related noise from meaningful scene features. This multi-domain processing enables the extraction of degradation-invariant features that remain consistent with those derived from clear images, thereby significantly enhancing model robustness under challenging weather conditions.}

To address the challenge of ineffective supervision, we aim to construct robust supervisory signals by introducing extra priors rather than solely relying on the photometric consistency assumption. 
Our inspiration stems from the pivotal assumption of \textit{scene correspondence}, which posits that a clear pair and its corresponding adverse pairs should exhibit the same semantic context and disparity.
However, capturing paired real-world outdoor datasets under diverse weather conditions in the same setting is challenging, and current stereo datasets lack sufficient training samples tailored to these diverse conditions.
To address this, we build on the outdoor DrivingStereo dataset~\cite{yang2019drivingstereo} and the Multi-Spectral Stereo dataset~\cite{shin2023deep} to generate paired adverse weather datasets.
Specifically, we employ the latest image-to-image translation model~\cite{parmar2024one} (CycleGAN-Turbo) to generate synthetic pairs for each clear stereo pair, creating realistic adverse weather scenes (i.e., rainy, foggy, night).  
These paired clear and adverse weather image pairs, which preserve the \textit{scene correspondence} property, are then used to construct reliable supervisory signals.



Benefiting from the \textit{scene correspondence} prior, we propose a two-step self-supervised pipeline trained on paired datasets to improve performance in challenging scenarios. 
This approach consists of two key steps: (1) self-supervised scene correspondence learning and (2) adverse weather distillation, both aimed at improving model disparity estimation under adverse conditions.
In the first step, we introduce two branches to process pairs of clear and adverse weather conditions, respectively.
Leveraging the \textit{scene correspondence} prior, we propose two consistency losses: Feature Consistency Loss and Disparity Consistency Loss.
These losses capitalize on the fact that the underlying scene maintains the same semantic information and disparity values between clear and adverse weather conditions. By ensuring consistency in both the learned features and disparity values across both branches, we provide robust supervisory signals that enable the model to learn a latent space less affected by weather degradations.
We then proceed with the adverse weather distillation step by using high-quality pseudo labels of clear pairs from the frozen model (Step 1) as supervisory signals, effectively mitigating the failures of the photometric consistency assumption in ill-posed regions, such as occluded areas.

Based on these preliminaries, we present RoSe, a \textbf{Ro}bust \textbf{Se}lf-supervised stereo matching method for adverse weather conditions.
We validate the effectiveness of our approach through extensive experiments on the DrivingStereo and Multi-Spectral Stereo (MS2) datasets, which encompass adverse weather conditions.
We create the Adverse-KITTI stereo datasets using the CycleGAN-turbo model.
When trained on this dataset, our method achieves highly competitive performance on both the standard KITTI 2012 and KITTI 2015 benchmarks~\cite{geiger2012we,menze2015object}.
These results show that the proposed framework largely outperforms previous self-supervised works under both standard and challenging conditions.
Our main contributions are summarized as follows.
\begin{itemize} 
\item
To our knowledge, we are the first to address self-supervised stereo matching across various adverse conditions. 
We propose RoSe, a robust self-supervised stereo matching pipeline for adverse weather conditions.
\end{itemize}


\begin{itemize}
    \item We propose a robust feature extractor, the Vision Foundation Model, coupled with a Feature Pyramid Network (FPN) and an Anti-Adverse Feature Enhancement Module (AFEM) for generating degradation-invariant features. This design aims to improve feature robustness against various adverse weather scenarios.
\end{itemize}

\begin{itemize}
    \item We present a two-step self-supervised pipeline that incorporates two consistency losses, which focus the model on learning resilience to weather effects, gradually improving performance under adverse weather conditions.
\end{itemize}

\begin{itemize} 
\item We conduct extensive experiments to validate the effectiveness of the proposed solution and show substantial performance improvements over existing State-of-the-Art (SoTA) self-supervised solutions.
\end{itemize}

\section{Related Works}
\subsection{Supervised Stereo Matching}
Current SoTA stereo matching networks can be roughly categorized as volume-based and volume-free.
For volume-based methods, some works~\cite{shen2021CFNet,shen2022pcw} propose a multi-scale cost volume strategy to narrow the disparity search space progressively.
Recent advances have further enhanced performance by introducing different cost volume aggregation strategies~\cite{wang2022spnet,guo2022cvcnet,wang2024cost,li2024inter,chen2023unambiguous,zeng2023deep,li2025globalregulationexcitationattention}. 
More recently, RAFT-Stereo~\cite{Lipson2021RAFTStereoMR}, GMStereo~\cite{xu2023unifying}, and Selective-IGEV~\cite{wang2024selective} adopt a cascaded recurrent network to update disparity iteratively.
For volume-free methods, CroCo-Stereo~\cite{weinzaepfel2023croco} uses an encoder-decoder transformer for cross-view completion and refines disparity by incorporating a dense prediction transformer head~\cite{ranftl2021vision}.
These supervised methods primarily address standard weather conditions, with few focusing on adverse conditions. 
Song \textit{et al.}~\cite{song2020simultaneous} integrates stereo matching and dehazing through a dual-branch structure with feature fusion. FoggyStereo~\cite{yao2022foggystereo} uses depth cues to create a foggy volume using estimated scattering coefficients for disparity estimation. CFDNet~\cite{liu2024cfdnet} employs feature contrastive learning to balance feature representation between clear and foggy domains. 
However, these methods rely on specialized modules for estimating scattering coefficients or feature fusion, which are not adaptable to other challenging scenarios.
Besides, all the above methods rely on Ground Truth, which is prohibitively expensive to acquire in large-scale real-world settings.
Therefore, self-supervised solutions are explored and have the potential to build foundation models in stereo matching by scaling data up to billions.

\subsection{Self-supervised Stereo Matching}
\label{sec2.1}
Self-supervised stereo methods bypass the need for Ground Truth by adopting a photometric consistency loss on the stereo pairs.
SssMnet~\cite{zhong2017ssl} proposes a loop photometric consistency loss.
Segstereo~\cite{yang2018segstereo} proposes to incorporate semantic cues to guide stereo matching.
Later, Flow2Stereo~\cite{liu2020flow2stereo} proposes a unified method to jointly learn optical flow and stereo matching.
OASM~\cite{li2021unsupervised} proposes an occlusion-aware stereo network to exploit occlusion cues as a depth cue for stereo matching.
PAM~\cite{wang2020parallax} proposes a parallax attention mechanism to capture the stereo correspondence without limiting disparity variations. 
ChiTransformer~\cite{su2022chitransformer} uses monocular cues and vision transformer~\cite{dosovitskiy2020image} (ViT) with cross-attention to enhance disparity estimation.
DualNet~\cite{wang2025dualnet} introduces negative-free contrastive learning into stereo matching by first training a teacher network through feature-metric consistency, and then using it to supervise the student’s probability distribution in a second stage.
Overall, existing self-supervised stereo matching methods face significant performance degradation under challenging conditions such as night, rain, and fog. This degradation occurs because these adverse conditions introduce noise into the pixel correspondences, undermining the photometric consistency assumption. Consequently, this poses substantial challenges for real-world applications, such as autonomous driving.

\textbf{Adverse Condition.}
In adverse weather conditions such as nighttime, rain, and fog, limited visibility and noise prevent the establishment of correct correspondences. 
So far, only several self-supervised works have explored disparity estimation under adverse weather in a self-supervised manner.
Sharma \textit{et al.}~\cite{sharma2018into} attempts to address nighttime stereo matching by optimizing a joint structure stereo model that performs structure extraction and stereo estimation jointly.
Subsequently, they~\cite{sharma2020nighttime} utilize GANs to render night stereo pairs from day stereo pairs, which are used to train a stereo network using the disparity supervision from the corresponding day pairs.
DTD~\cite{vankadari2024dtd} uses a masking method and distance regularizer to enhance the accuracy of depth estimation at nighttime.
However, these works only consider a single adverse modality and ignore that weather has varying categories, which hinders the application of the stereo model under adverse conditions.
Unlike these works, we propose a simple and effective solution that enables a stereo model in a self-supervised manner to reliably estimate disparity in various challenging conditions (e.g., night, fog, and rain) with a few burdens. 

\subsection{Feature Representation Learning in Stereo}
Feature representation learning is paramount in achieving robust performance by accurately identifying pixel-wise correspondences between rectified stereo pairs.
As a common feature extraction strategy in stereo matching, Feature Pyramid Network (FPN) with different dilation settings is widely adopted~\cite{shen2021CFNet,wang2025adstereo,zhang2025lgast}. 
Then, a modified multi-scale feature extraction mechanism~\cite{shen2021CFNet,shen2022pcw} is proposed to construct a fused cost volume representation, improving the robustness of the model.
Recently, Vision Foundation Models (VFMs), such as SegmentAnything (SAM)~\cite{kirillov2023segment} and DepthAnything~\cite{yang2024depth,depth_anything_v2} have rapidly become a hot topic in the field of computer vision.
Zhang \emph{et al.}~\cite{zhanglearning2024,zhang2025panmatch} adapt the pre-trained VFMs by proposing a generalized feature adapter and well-designed modules and functions to obtain domain-invariant features, substantially improving robustness. However, they might still face challenges in extracting fine-grained feature representations specific to the target domain from coarse-grained features of the pre-trained VFM, which could result in suboptimal performance~\cite{goyal2023finetune}.
Wang \emph{et al.}~\cite{wang2025learning} propose a selective Mixture-of-Experts strategy in the pre-trained ViT block to flexibly adapt the feature representation.
Unlike them, we explore a simple way by integrating the powerful Vision Foundation Model (VFM) with FPN for feature extraction, empowering the model with discriminative and robust features across various scenarios.

\subsection{Knowledge Distillation for Depth Estimation}

Knowledge distillation (KD) has been widely used in supervised monocular depth estimation \cite{Hu2022DenseDD,hutnnls2023}, self-supervised monocular depth estimation \cite{petrovai2022exploiting,liu2023self}, and stereo matching \cite{2023UCFNet}. 
Unlike these works, we leverage KD to address low robustness in adverse weather conditions.
While a few studies \cite{gasperini2023robust} tackle this issue in self-supervised monocular depth estimation using only unaugmented-augmented depth correspondences, our approach differs in three key ways: 1) integrating a foundation model as a stronger feature extractor, 2) introducing an anti-adverse feature enhancement module, and 3) regularizing both output and intermediate feature consistency. These innovations allow our method to achieve notable accuracy gains over previous SoTA methods.


\begin{figure*}[t]
    \includegraphics[width=1.0\linewidth]{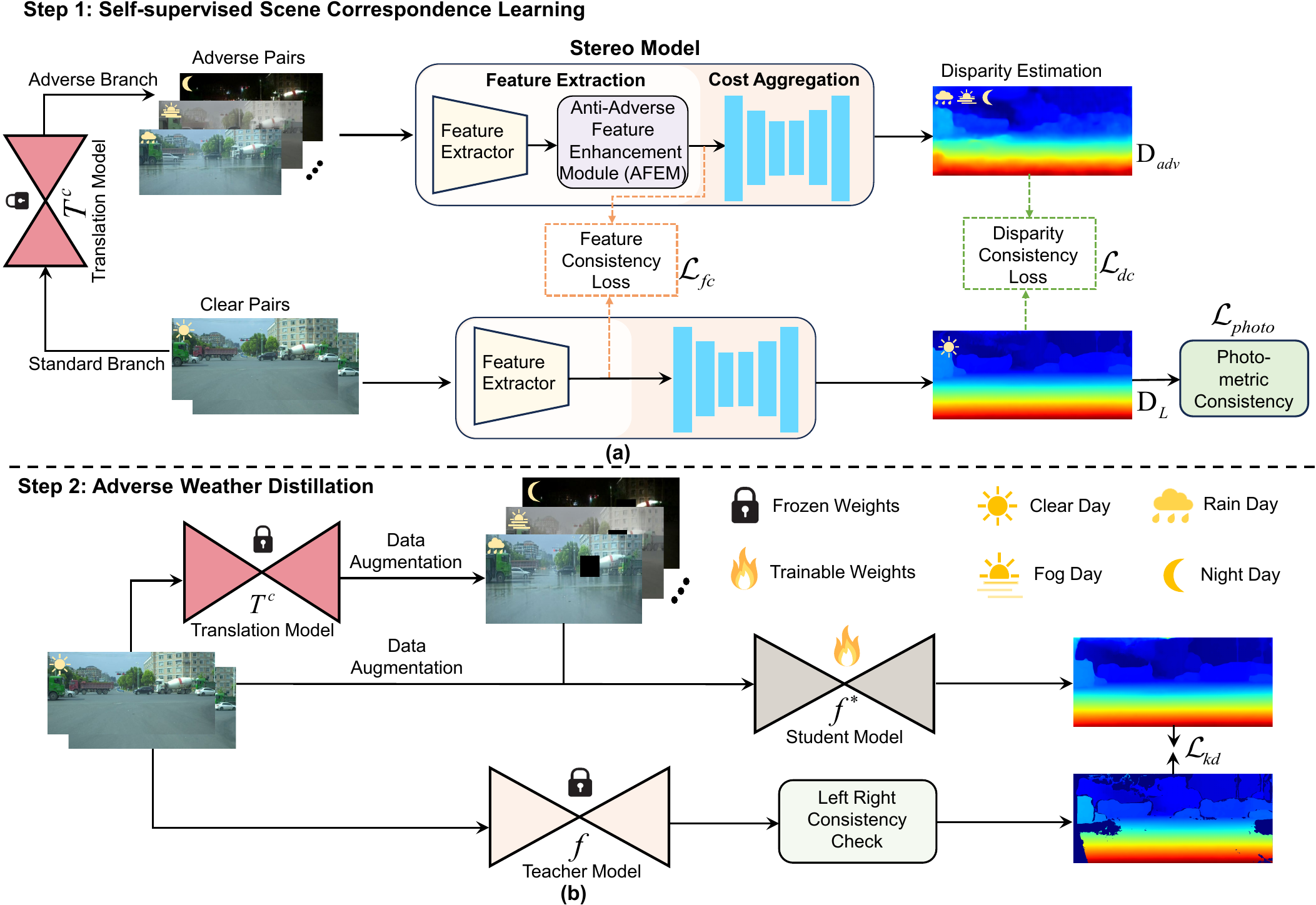}
    \vspace{-0.8cm}
    \caption{An overview of {RoSe}. 
    (a) denotes the self-supervised scene correspondence learning. Both branches share weights except for the feature extractors.
    In Step 2, the frozen stereo model from the first step acts as the teacher model, generating high-quality pseudo labels on clear samples and guiding the student model (trainable) with mixed clear and adverse inputs. Both the teacher and student models share the same architecture.}
    \label{sec3:step1}
    \vspace{-0.4cm}
\end{figure*}

\section{Methodology}

In this section, we present the robust self-supervised stereo pipeline, {RoSe}, which comprises two steps: self-supervised scene correspondence learning (Step 1) and adverse weather distillation (Step 2). The proposed self-supervised model shares the same architecture as RAFTStereo~\cite{Lipson2021RAFTStereoMR}, except for the feature extractor.
Unlike previous works that employ specialized structures~\cite{su2022chitransformer,wang2020parallax,yao2022foggystereo}, our method focuses on the training process and the feature extractor, enabling seamless integration into various stereo backbones. A schematic of the pipeline is illustrated in Fig.~\ref{sec3:step1}.

For Step 1, we introduce a self-supervised learning mechanism that leverages scene correspondence priors and robust priors from VFMs to compute consistency losses between clear and adverse pairs, serving as extra supervisory signals.
In Step 2, we take adverse weather distillation by using high-quality pseudo labels generated from the frozen model (Step 1) as supervisory signals. 
This approach effectively eliminates the reliance on photometric consistency loss, enhancing the model's robustness in challenging scenarios. Once trained, the final model (Step 2) is used for inference.

\vspace{-0.2cm}
\subsection{Preliminary}
\label{sec3:Preliminary}
Given a rectified stereo image pair ${\bf{I}}_{L}$, ${\bf{I}}_{R}$, 
the network first uses the modified feature extractor to extract features from stereo image pairs. 
Then, the left and right features are formed into a cost space, which is processed to predict the resultant disparity maps ${\bf{D}}_{L}$.
In self-supervised stereo matching, instead of relying on expensive ground truth labels, we use the input itself to supervise our model. 
The motivation is that if we can accurately warp between the image pairs, we must have learned the dense disparity map.
Correspondingly, the left image ${\bf{I}}_{L}$ can be reconstructed from the right image ${\bf{I}}_{R}$ as follows:
\begin{equation}
     {\bf\hat{I}}_{L}(i,j) = {\bf{I}}_{R}(i+{\bf{D}}_{L}(i,j),j), 
     \label{warp_loss}
\end{equation}
where (\textit{i, j}) represent the pixel coordinates and ${\bf\hat{I}}_{L}$ is the reconstructed left image. 
The hypothesis of photometric consistency is to maximize the similarity between the left image ${\bf{I}}_{L}$ and the reconstructed left image ${\bf\hat{I}}_{L}$ after being warped from the right perspective.
Correspondingly, the photometric consistency loss can be formulated as:
\begin{equation}
    \begin{split}
    \mathcal{L}_{photo} = \frac{1}{N}\sum_{i, j} &(\alpha\frac{1-\mathcal{S}({\bf{I}}_{L}(i,j),{\bf\hat{I}}_{L}(i,j))}{2} + \\
    &(1-\alpha)||{\bf{I}}_{L}(i,j)-{\bf\hat{I}}_{L}(i,j)||),
    \end{split}
     \label{photo_loss}
\end{equation}
where $N$ is the number of pixels, $\mathcal{S}$ is an SSIM function~\cite{wang2004image} and hyper-parameter $\alpha$ is emprically set to 0.85.
Besides, following ~\cite{wang2020parallax,ye2021unsupervised}, we use an edge-aware smoothness loss $\mathcal{L}_{s}$ to encourage local smoothness of the disparity:
\begin{equation}
\begin{split}
\mathcal{L}_{s}=\frac{1}{N}\sum_{i,j}&(||\nabla_{x}\textbf{D}_{L}(i,j)||_{1}\exp^{(-||\nabla_{x}\textbf{I}_{L}(i,j)||_{1})} + \\
&||\nabla_{y}\textbf{D}_{L}(i,j)||_{1}\exp^{(-||\nabla_{y}\textbf{I}_{L}(i,j)||_{1})}),
\end{split}
\end{equation}
where $\nabla_{x}$ and $\nabla_{y}$ are gradients in the x and y axes, respectively.
By minimizing the smoothness loss, the disparity map can be aligned with the edge structure of the input image while ensuring the smoothness of the disparity map.

\begin{figure*}[t]
    \includegraphics[width=1.0\linewidth]{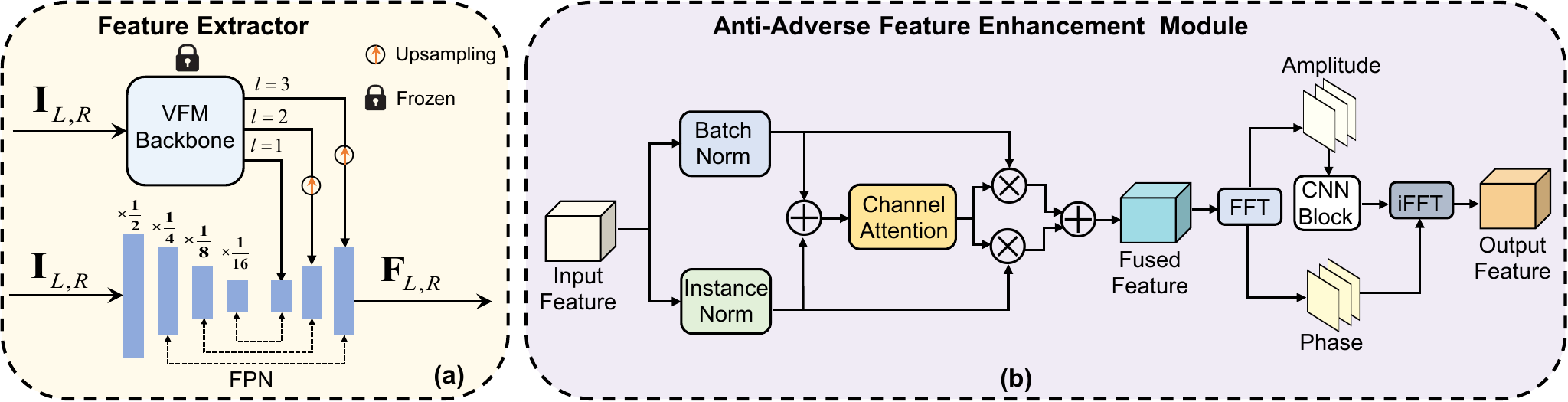}
    \vspace{-0.6cm}
    \caption{(a), an overview of feature extractor. (b),    the features at varied depths from the frozen VFM are integrated into the FPN for robust feature extraction. 
    (c), the Anti-Adverse Feature Enhancement Module (AFEM) is added to the adverse branch to generate degradation-invariant features.}
    \label{sec3:step2}
    \vspace{-0.2cm}
\end{figure*}

\subsection{Feature Extractors}
\label{sec3:Feature}
Current CNN-based feature extractors struggle to learn discriminative features from reflective and texture-less regions due to limited receptive fields and weak supervisory signals. 
Therefore, we integrate the pre-trained VFM (\textit{i.e.,} DAMV2~\cite{depth_anything_v2}) into FPN for robust feature extraction. 
The former aims to obtain robust and all-purpose scene priors, while the latter is devoted to capturing rich details on the target domain.

As shown in Fig.~\ref{sec3:step2} (a), we use two parallel branches to extract features from left and right images, and the final output is denoted as ${\bf{F}}_{L,R}$.
Specifically, before taking the stereo image pair to the pre-trained VFM, we resize the position embedding of the pre-trained VFM with bicubic interpolation to fit different image scales~\cite{dosovitskiy2020image}. 
Subsequently, features derived from vision transformation blocks at varying depths are collected. 
These aggregated features at low resolution (\textit{i.e.,} 1/16 of the original image resolution) are then subjected to bilinear interpolation operation to yield feature outputs at three diverse scales (${1}/{16}$, ${1}/{8}$, and ${1}/{4}$ of the original image resolution).
Next, these features at various scales are directly concatenated to those from an FPN encoder, and a regular FPN decoder is employed to produce the final feature outputs ${\bf{F}}_{L}$, ${\bf{F}}_{R}$.
These features contain rich content information from both ViT and FPN, and will be used to formulate an informative and discriminative correlation pyramid~\cite{Lipson2021RAFTStereoMR}. 

\textbf{Anti-Adverse Feature Enhancement.}
Although the proposed feature extractor can yield discriminative features for clear pairs, adverse weather conditions compromise the performance of robust priors in VFMs, leading to a deterioration in the quality of the learned features~\cite{qiao2023robustness,shan2023robustness,wang2024empirical}.
To mitigate this issue, we propose an Anti-Adverse Feature Enhancement Module (AFEM). This module is designed to extract degradation-invariant feature representations that are consistent with those obtained from clear images by the original feature extractor. 
Consequently, AFEM improves the feature robustness across various challenging scenarios.

As illustrated in Fig.~\ref{sec3:step2} (b), the input features are initially processed using Instance Normalization (IN) and Batch Normalization (BN) in parallel. IN standardizes variations associated with image degradation, individually removing style attributes (such as adverse weather styles) while preserving core content~\cite{huang2017arbitrary,son2020urie,zhang2024artbank}. 
This is essential to mitigate the influence of image distortions and ensure content stability under various adverse weather conditions.
Conversely, BN is trained by considering various degraded images together within each mini-batch, which can maintain the distortion information disregarded by the IN process~\cite{son2020urie}.
Consequently, these two normalization processes complement each other and improve the understanding of input weather degradations.


Then, we use a channel attention mechanism to scrutinize the merged features,  generating attention maps for each channel. This process dynamically weighs the importance of each feature type, resulting in a feature set that leverages the advantages of both normalization techniques~\cite{son2020urie}.
Moreover, inspired by~\cite{xu2021fourier,yang2020fda,huang2022deep}, we introduce a Fourier Degradation Suppression module to enhance the integrated features by transforming them from the spatial to the frequency domain using the Fast Fourier Transform (FFT). Given that the amplitude components in the FFT represent style information related to image degradation, we aim to isolate and filter these degradation elements by applying a shallow 2D CNN block, while preserving the phase components to maintain structural integrity. The refined features are then returned to the spatial domain via inverse FFT (iFFT). This entire process treats adverse weather degradations as image styles, processing them across the spatial, channel, and frequency domains, generating degradation-invariant features for robust stereo matching.

\subsection{Self-supervised Scene Correspondence Learning}
When training directly on adverse weather data using the naive losses outlined in Sec.~\ref{sec1:intro}, there is performance degradation compared to that of clear inputs.
This is because the photometric loss is vulnerable to illumination variations and noise patterns (i.e., texture-less, blur, and reflections), which disrupt pixel correspondence across views, as revealed in Sec.~\ref{sec4:ablation_study}.
To mitigate this, we build on our designs from the key observation:
The generated synthetic pairs under adverse weather conditions retain the same scene information as their clear counterparts. 
This motivates us to leverage a paired-data training pipeline to construct valid supervisory signals. These supervisory signals can enhance the model's robustness under adverse conditions, similar to the mechanisms observed in contrastive learning~\cite {chen2020simple} (clear samples \textit{vs.} adverse samples).

\textbf{Consistency Losses.}
Based on this key observation, we first leverage the image-to-image translation model (CycleGAN-Turbo)~\cite{parmar2024one} to generate synthetic pairs with adverse weather styles, ensuring that their scene information can be consistent with that of clear ones.
Next, the adverse pair is first fed into the adverse branch to produce the stereo features ${\bf{F}}^{adv}_{L,R}$ (denoted as left and right image features) and infer the adverse disparity map ${\bf{D}}^{adv}$, as illustrated in the top half of Fig.~\ref{sec3:step1} (a). 
Similarly, the predictions of the clear pair are denoted as ${\bf{F}}^{clr}_{L,R}$ and ${\bf{D}}^{clr}$ (bottom half in Fig.~\ref{sec3:step1} (a)). 
By leveraging such \textit{scene correspondence}, we propose two losses to improve the model’s resilience to adverse weather conditions: the feature consistency loss and the output consistency loss. Both losses take advantage of the paired clear and adverse images in the used datasets.

To improve the robustness of our feature extractor when facing challenging conditions, we impose a feature consistency loss $\mathcal{L}_{fc}$. This loss minimizes the discrepancy between the features learned under clear and adverse weather conditions, ensuring consistent feature representation across varying degraded scenarios:
\begin{equation}
    \mathcal{L}_{fc} = 1 - \frac{\mathbf{F}^{adv}_{L,R}\cdot{\mathbf{F}^{clr}_{L,R}}}{\Vert\mathbf{F}^{adv}_{L,R}\Vert\Vert\mathbf{F}^{clr}_{L,R}\Vert} ,
    \label{eq:fc}
\end{equation}
where $\mathcal{L}_{fc}$ is the feature consistency loss. 
This process aligns the learned features from both clear and adverse pairs in the latent space.
By minimizing the feature consistency loss, we ensure that the refined features in adverse conditions remain consistent with those extracted under clear image conditions, thus guaranteeing the robustness and consistency of the features across different weather degradations.

Following~\cite{Lipson2021RAFTStereoMR}, we additionally propose a disparity consistency loss to minimize the following:
\vspace{-0.1cm}
\begin{equation}
\mathcal{L}_{dc}=\sum^{N}_{i=1}\beta^{N-i}||({\bf D}^{clr}-{\bf D}^{adv}_{i}) \odot {\bf M}_{v})||_{1},
\end{equation}
where $N = 16$ denotes the number of iterations during training, and the exponential weight $\beta$ is set to 0.9. $\odot$ is an element-wise product and the binary valid mask $\textbf{M}_{v}$ is used to filter the unreliable point.

\textbf{Geometric Confidence Mask.}
We define the per-pixel geometric confidence mask $\textbf{M}_{v}$ by employing the widely used left-right consistency check~\cite{hirschmuller2007stereo} to filter outliers. 
For any arbitrary point $\textbf{\textit{p}}(i,\ j)$ for the left image, if its disparity in the left disparity map ${\bf D}_{L}$ is $d_{l}$, then the disparity value of the corresponding point $\textbf{\textit{p}}(i-d_{l},\ j)$ in the right disparity map ${\bf D}_{R}$ is denoted by $d_{r}$.
Thus, the warp error at point $\textbf{\textit{p}}$ can be written as $e_{lr}=||d_{l}-d_{r}||$. $\textbf{M}_{v}$ can be obtained as follows:

\begin{equation}
    {\bf M}_{v} = \begin{cases}
	1, &\; 
    e_{lr} \leq \tau \\
	0, &\; otherwise\\
    \end{cases},
    \label{sec3:lrc}
\end{equation}
where $\tau$ is the threshold and is set to 1 pixel in our settings.
Except for the disparity consistency loss $\mathcal{L}_{dc}$, ${\bf M}_{v}$ is also used to regularize the following distillation loss $\mathcal{L}_{kd}$.

To summarize, the final loss function for the self-supervised scene correspondence learning mechanism is:
\begin{equation}
    \mathcal{L}_{self} = \lambda_{1}\mathcal{L}_{photo} + \lambda_{2}\mathcal{L}_{s} + \lambda_{3}\mathcal{L}_{fc} + \lambda_{4}\mathcal{L}_{dc},
\end{equation}
where $\lambda_{i}$ is a hyperparameter and we empirically set \revise{$\lambda_{1}$ = $\lambda_{3}$ = $\lambda_{4}$ = 1, and $\lambda_{2}$ = 10.} 
Please note that only adopting $\mathcal{L}_{fc}$ and $\mathcal{L}_{dc}$ is not feasible, as it can lead to training collapses with the learned features and disparity maps converging to constant values (typically 0)~\cite{ding2022kd}.

\subsection{Adverse Weather Distillation}
Since our method relies on the photometric consistency assumption in Step 1, performance can be affected by ill-posed areas such as occluded regions. 
To address this, we enhance our {RoSe} (Step 1) by using high-quality pseudo-supervised loss to replace the photometric consistency loss, providing robust supervisory signals regardless of the adverse conditions of the input.
The motivation for this approach stems from the empirical success of deep stereo models, which have shown robust performance even when trained with sparse ground truth labels or noisy estimates~\cite{2023UCFNet}. For example, deep models have achieved impressive results on the KITTI 2015 dataset, despite having less than 30\% of the ground truth annotated for 200 training images.

\revise{Specifically, we employ an adverse weather distillation strategy. First, a frozen teacher model, finetuned in Step 1, generates pseudo disparity maps from clear image pairs (as illustrated in Fig.~\ref{sec3:step1} (b)). To ensure the quality of these pseudo labels, we apply the Geometric Confidence Mask $\textbf{M}_{v}$ to filter out unreliable regions.}

\revise{The student model is then trained using these high-quality pseudo labels as supervision. Its input consists of pre-processed mixed pairs (clear and adverse), which are enhanced with data augmentations like asymmetric occlusions. This training regimen simulates photometric consistency failures, teaching the student to infer disparity maps under more challenging conditions. This entire distillation mechanism enhances the model's reliability and performance, and is governed by the following knowledge distillation loss:}
\begin{equation}
    \mathcal{L}_{kd} = \frac{1}{N} \sum_{p=1}^{N} |(f^{*}(m_{i})_{p}-f(c_{i})_{p})\odot {\bf M}_{v}(p)|.
\end{equation}
Where $N$ is the number of pixels, $f^{*}(m_{i})$ is the being trained stereo network's disparity predictions on a mixed sample pair $m_{i}$ (i.e., a clear or adverse pair), and $f(c_{i})$ is the disparity estimation of the frozen stereo network $f$ on a clear pair $c_{i}$.
$f^{*}$ aims to learn to follow $f$ at the output level without being affected by adverse inputs.

\section{Experiments}
In this section, we first introduce the datasets and metrics, Vision Foundation Models, and implementation details employed in our experiments. Following this, we conduct a detailed comparison of zero-shot performance against recent state-of-the-art (SoTA) stereo matching methods. 
Subsequently, we compare our approach with previous self-supervised SoTA methods across these three test sets (DrivingStereo, MS2, and KITTI), as detailed in Sec.~\ref{sec4:sota}. 
Finally, we present comprehensive ablation studies to demonstrate the effectiveness of our network, as discussed in Sec.~\ref{sec4:ablation_study}.



\begin{table}[t]
\centering
\scriptsize
\caption{We evaluate the domain generalization ability with checkpoints provided by authors.  Since the Drivingstereo dataset lacks night modality, instead, we adopt the night test set in MS2. All models are pre-trained using only SceneFlow in a supervised manner. Bad 3.0 metric is adopted  (Best results
in \textbf{bold} and sub-optimal best in \textbf{\textcolor{blue}{blue}}).}
\vspace{-0.2cm}
\setlength{\tabcolsep}{1.5mm}{
    \begin{tabular}{c|ccc|c|c}
    \toprule
    \multirow{2}{*}[-1pt]{\textbf{Method}} & \multicolumn{3}{c|}{\textbf{Drivingstereo}} & {\textbf{MS2}} &  \multirow{2}{*}[-1pt]{\textbf{Avg.}} \\
    &{\textbf{Clear (Sunny)}}$\downarrow$  & {\textbf{Foggy}}$\downarrow$ &{\textbf{Rainy}$\downarrow$} & {\textbf{Night}}$\downarrow$ \\
    \midrule
    GwcNet~\cite{guo2019group} & 9.29 & 15.8 & 16.5 & 34.3 &  19.0\\
    CFNet~\cite{shen2021CFNet} & {5.35} & 6.39 & \textbf{\textcolor{blue}{11.4}} & 33.4 & 14.1 \\
    GMStereo~\cite{xu2023unifying} & 14.6 & 16.6 & 23.9 & 46.5 & 25.3\\
    IGEV~\cite{xu2023iterative} & 5.53 & {4.96} & 15.7 & \textbf{\textcolor{blue}{29.2}} & \textbf{\textcolor{blue}{13.9}} \\
    Selective-IGEV~\cite{wang2024selective} & 6.03 & \textbf{\textcolor{blue}{4.98}} & 14.3 & 30.5 & 14.0\\
    Former-PSMNet~\cite{zhanglearning2024} & \textbf{3.5} & 6.3 & 12.7 & - & - \\
    \midrule
    \textbf{RoSe (Pre-trained)} & \textbf{\textcolor{blue}{3.63}} & \textbf{4.85} & \textbf{4.75} & \textbf{20.8} &  \textbf{8.5}\\
    \bottomrule
    \end{tabular}}
\label{sec4:dg}
\vspace{-0.3cm}
\end{table}

\subsection{Datasets and Metrics.}

\textbf{SceneFlow.}
SceneFlow~\cite{mayer2016large} is a large-scale synthetic dataset with 35454 training images and 4370 test images of $960 \times 540$. This dataset provides dense ground truth disparity maps for stereo matching. 
We use the synthetic dataset to pre-train our model and evaluate its generalization ability.
The maximum disparity range is set to 192.

\textbf{DrivingStereo.} 
DrivingStereo~\cite{yang2019drivingstereo} is an outdoor stereo dataset in driving scenarios.
Since the training set contains scenes with rain and fog that may violate the photometric consistency assumption, we exclude these scenes and only use the good visibility day sets with 84665 stereo pairs for training. 
We use the official weather split (500 clear pairs, 500 cloudy pairs, 500 foggy pairs, and 500 rainy pairs) to evaluate the performance of our models in real-world conditions.
Note that, we find that the cloudy weather set with good visibility, which is identical to the clear pairs. As a result, we focus our evaluation solely on the clear, foggy, and rainy weather validation sets to ensure diverse and challenging conditions are adequately represented in our assessments.
Since this dataset does not contain night conditions, we generate test night stereo pairs using the CycleGAN-Turbo model~\cite{parmar2024one} based on its test set (7751 pairs).
The maximum disparity range is set to 128.

\textbf{MS2.} 
The MS2 dataset~\cite{shin2023deep} is taken from multi-spectral sensors on a vehicle in Daejeon, South Korea.
This dataset provides different weather conditions (clear, rainy, and night).
Following the officially provided training split, we exclude night and rainy scenes and only use the good visibility (clear) day sets with 70659 stereo pairs for training. 
We use the official weather split (23317 clear pairs, 25023 rainy pairs, and 18389 night pairs) to evaluate the performance of our models in real-world conditions.
Since this dataset does not contain foggy scenes, we generate test foggy scenes using the same image translation model~\cite{parmar2024one} based on the clear test set (23317 pairs).
The maximum disparity range is set to 128.

\begin{table}[t]
\centering
\scriptsize
\caption{Domain generalization evaluation on target training sets. $^\ddag$ denotes the ViT-Large capacity. $\ast$ denotes that we use the officially provided weights to evaluate.}
\vspace{-0.1cm}
\setlength{\tabcolsep}{.2mm}{
    \begin{tabular}{l|cccc}
    \toprule
    \multirow{2}{*}[-1pt]{\; Method} & {\textbf{KITTI 2012}} & {\textbf{KITTI 2015}}  & {\textbf{Middle}} & \textbf{ETH3D}\\
    & \textbf{Bad 3.0 (\%)} & \textbf{Bad 3.0 (\%)}  & \textbf{Bad 2.0 (\%)} &  \textbf{Bad 1.0 (\%)} \\
    \midrule
    GwcNet~\cite{guo2019group} & 20.2 & 22.7 & 34.2 & 30.1 \\
    DSMNet~\cite{zhang2020domain} & 6.2 & 6.5 & 14.1 & 6.2 \\
    CFNet~\cite{shen2021CFNet} & 4.7 & 5.8 & 15.4 & 5.8 \\
    Mask-CFNet~\cite{rao2023masked} & 4.8 & 5.8 & 13.7 & 5.7 \\
    PCWNet~\cite{shen2022pcw} & \textbf{4.2} & 5.6 & 15.8 & ${5.2}$ \\
    RAFT-Stereo~\cite{Lipson2021RAFTStereoMR} & 5.1 & 5.7 & 12.6 & \textbf{3.3}\\
    CREStereo~\cite{li2022practical}  & 6.7 & 6.7 & 15.3 & 5.5 \\
    UCFNet\_pretrain~\cite{2023UCFNet}  & 4.5 & 5.2 & 26.0 & 4.8 \\
    Former-PSMNet$^\ddag$ (SAM)~\cite{zhanglearning2024}& $\textbf{\textcolor{blue}{4.3}}$ & $\textbf{5.0}$ & $\textbf{9.4}$ & 6.4 \\
    \midrule
    \textbf{RoSe (Pre-trained)} & $\textcolor{blue}{\textbf{4.3}}$  & $\textcolor{blue}{\textbf{5.1}}$ &  $\textbf{\textcolor{blue}{12.2}}$  & $\textcolor{blue}{\textbf{3.4}}$\\
    \bottomrule
    \end{tabular}}
    \vspace{-0.3cm}
\label{sec4:dg_comparison}
\end{table}

\begin{table*}[t]
    \footnotesize
    \centering    
    \caption{Evaluation of self-supervised stereo methods on the DrivingStereo validation weather set. 
    Sup. indicates whether the method is supervised or not. 
    We evaluate the efficiency metrics of these methods using the same hardware architecture (NVIDIA RTX A100).
    Training data (Tr.data): d: day-clear, T: translated in adverse weather, n: nighttime, f: foggy day, r: rainy day. (The self-supervised best results in \textbf{bold} and the sub-optimal best results in \color{blue}{\textbf{blue}}).}
    \vspace{-0.2cm}
    \setlength{\tabcolsep}{2mm}{
    \begin{tabular}{c|ccc|cc|cc|cc|cc|cc}
        \toprule
         \multirow{2}{*}[-1pt]{\textbf{ID}}  &\multirow{2}{*}[-1pt]{\textbf{Method}}  & \multirow{2}{*}[-1pt]{\textbf{Sup.}} & \multirow{2}{*}[-1pt]{\textbf{Tr.Data}} & \multicolumn{2}{c|}{\textbf{Clear}}  & \multicolumn{2}{c|}{\textbf{Foggy}} & \multicolumn{2}{c|}{\textbf{Rainy}} & \multicolumn{2}{c}{\textbf{Night (Synthetic)}} & \multicolumn{1}{c}{\textbf{Mem.}}  & \multicolumn{1}{c}{\textbf{Time}} \\
        &&&& \textbf{EPE}$\downarrow$ & \textbf{Bad 3.0} $\downarrow$ & \textbf{EPE} $\downarrow$& \textbf{Bad 3.0} $\downarrow$ &\textbf{EPE} $\downarrow$ & \textbf{Bad 3.0} $\downarrow$ & \textbf{EPE}$\downarrow$ & \textbf{Bad 3.0} $\downarrow$ & \textbf{(GB)} & \textbf{(s)} \\
         \midrule
         1 & SGM~\cite{hirschmuller2007stereo} & - & - & 1.08 & 7.75 & 1.89 & 12.7 & 3.00 & 21.8 & 1.46 & 11.0 & - & 79 \\
         \midrule
         2 &  PASMNet~\cite{wang2020parallax} & \XSolidBrush & $\textit{\textbf{d}}$ & 1.44 & 6.11 & 1.53 & 7.20 & 2.77 & 16.2 & 3.67 & 23.1 & 1.9 & 0.08 \\
         3 &  PASMNet~\cite{wang2020parallax} & \XSolidBrush & $\textit{\textbf{dT(nfr)}}$ & 1.62 & 7.01  & 1.85 & 99.48 & 3.11 & 16.9 & 9.89 & 40.9 & 1.9 & 0.08 \\
         4 &  ChiT~\cite{su2022chitransformer} & \XSolidBrush & $\textit{\textbf{d}}$ & 1.01 & 4.84 & 1.26 & 5.93 & 2.24 & 13.1 & {1.56} & {11.7} & 4.7 & 0.17 \\
         5 &  ChiT~\cite{su2022chitransformer} & \XSolidBrush &  $\textit{\textbf{dT(nfr)}}$ & 1.17 & 5.35 & 1.49 & 6.87 & 2.60 & 15.6 & 4.41 & 26.8 & 4.7 & 0.17 \\
         6 & DTD~\cite{sharma2020nighttime} & \XSolidBrush &  $\textit{\textbf{T(n)}}$ & - & - & - & - & - & - & \textbf{\textcolor{blue}{1.21}} & \textbf{\textcolor{blue}{7.41}} & 7.9 & 1.1 \\
         \midrule
        7 &  \textbf{Rose} & \XSolidBrush & $\textit{\textbf{d}}$ & \textbf{0.74}  & \textbf{1.31} & \textbf{\textcolor{blue}{1.07}} & \textbf{\textcolor{blue}{3.76}} & \textbf{\textcolor{blue}{1.47}} & \textbf{\textcolor{blue}{7.67}} & 1.93 & 12.9  & 2.7 & 0.13 \\
        8 & \textbf{Rose} & \XSolidBrush & $\textit{\textbf{dT(nfr)}}$ & \textbf{\textcolor{blue}{0.78}} & \textbf{\textcolor{blue}{1.60}} & \textbf{0.85} & \textbf{1.71} & \textbf{0.88} & \textbf{1.88} & \textbf{1.01} & \textbf{3.14} & 2.7 & 0.13 \\
         \bottomrule
    \end{tabular}}
    \label{tab:dr}
    \vspace{-0.2cm}
\end{table*}

\textbf{KITTI 2012 \& 2015.} 
KITTI 2012~\cite{geiger2012we} contains outdoor driving scenes with sparse ground truth disparities. It contains 194 training samples and 195 testing samples with a resolution of $370\times1226$.
KITTI 2015~\cite{menze2015object} contains driving scenes with sparse disparity maps. It contains 200 training samples and 200 test samples with a resolution of $375\times1242$.
Since the KITTI dataset only contains clear weather mode, we generate foggy, rainy, and night weather stereo pairs using the CycleGAN-turbo model based on the training sets (394 pairs).
We use a mix of KITTI 2012 \& 2015 adverse weather training sets for training ($394\times 4$ image pairs). 
We evaluate its test set (clear weather) to compare our model with previous self-supervised works under standard conditions.
The maximum disparity range is set to 192.

\textbf{Synthetic Datasets.}
For these clear training datasets (i.e., DrivingStereo, MS2, KITTI), 
we use the CycleGAN-Turbo model $T
^{c}$~\cite{parmar2024one} to translate each clear stereo pairs with good visibility ($c_{i}$) into different adverse conditions ($h_{i}^{c}$), with $c\in C$ = \{night, rain, fog\}.
For day-to-night translation, we use random $512 \times 512$ crops from BDD100K~\cite{yu2020bdd100k} for training the CycleGAN-Turbo model.
For day-to-fog translation, we use random $512 \times 512$ crops from DENSE~\cite{bijelic2020seeing} and FoggyCityScapes~\cite{sakaridis2018semantic}  datasets for training the CycleGAN-Turbo model.
For day-to-rain translation, we use random $512 \times 512$ crops from Nuscenes~\cite{caesar2020nuscenes} for training.
Leveraging the CycleGAN-Turbo model, we develop three comprehensive datasets to enhance our model's robustness in diverse weather conditions:
1) The Adverse-DrivingStereo dataset features $84665 \times 4$ stereo image pairs across four weather conditions: clear, foggy, rainy, and night. 
2) The Adverse-MS2 dataset includes $79659 \times 4$ stereo image pairs under the same conditions. 
3) Additionally, the Adverse-KITTI dataset comprises $394 \times 4$ stereo image pairs in these varied weather scenarios. We use these three extensive synthetic datasets to train our model in a pairwise manner (clear \textit{vs.} adverse weather), ensuring robust performance across diverse and challenging scenarios.

\textbf{Evaluation Metric.}
For evaluation metrics, end-point error (EPE) and t-pixel error rate ( Bad t) are adopted.
In addition, the percentage of stereo disparity outliers, defined as disparities larger than 3 pixels or more than 5\% of the ground truth disparities (referred to as D1), is utilized as the metric.

\subsection{Vision Foundation Models}
\textbf{SAM.} 
SegmentAnything~\cite{kirillov2023segment} is a robust segment segmentation model that has demonstrated its extensive applicability to various dense prediction tasks, showcasing its versatility and effectiveness.
We employ its ViT-Base architecture as our image encoder, making use of pre-trained weights that were trained on SA-1B~\cite{kirillov2023segment}. The patch size of this model is set to 16×16, and each layer is designed to produce features with a dimensionality of 768, summing up to a total of 12 layers (ViT-Base).
The positional embeddings of the model are upscaled to a length of 1024 via bicubic interpolation. From this model (ViT-Base), we extract features from the 5th, 7th, and 11th layers and feed them into the FPN decoder.

\textbf{DAMV2.}
DepthAnythingV2~\cite{depth_anything_v2} designed a data engine to automatically generate depth annotations for unlabeled images, enabling data scaling up to an arbitrary scale.
Similar to SAM, we employ the ViT-Base architecture as our image encoder.
The patch size of this model is set to 16×16, summing up to a total of 12 layers.
The positional embeddings of the model are upscaled to a length of 1024 via bicubic interpolation. From this model, we extract features from the 5th, 7th, and 11th layers and feed them into the FPN decoder.


\begin{table*}[t]
    \footnotesize
    \centering
    \caption{Evaluation of self-supervised stereo methods on the MS2 validation weather set.}
    \vspace{-0.2cm}
    \setlength{\tabcolsep}{2.8mm}{
    \begin{tabular}{c|ccc|cc|cc|cc|cc}
        \toprule
        \multirow{2}{*}[-1pt]{\textbf{ID}}  & \multirow{2}{*}[-1pt]{\textbf{Method}}  & \multirow{2}{*}[-1pt]{\textbf{Sup.}} & \multirow{2}{*}[-1pt]{\textbf{Tr.Data}} & \multicolumn{2}{c|}{\textbf{Clear}}  & \multicolumn{2}{c|}{\textbf{Foggy (Synthetic)}} & \multicolumn{2}{c|}{\textbf{Rainy}} & \multicolumn{2}{c}{\textbf{Night}} \\
        &&&& \textbf{EPE}$\downarrow$ & \textbf{Bad 3.0} $\downarrow$ & \textbf{EPE} $\downarrow$ & \textbf{Bad 3.0} $\downarrow$ &\textbf{EPE} $\downarrow$ & \textbf{Bad 3.0} $\downarrow$ & \textbf{EPE}$\downarrow$ & \textbf{Bad 3.0} $\downarrow$\\
         \midrule
         1 & SGM~\cite{hirschmuller2007stereo} & - & - & 1.11 & 9.69 & \textbf{\textcolor{blue}{1.23}} & 11.8 & \textbf{\textcolor{blue}{1.17}} & \textbf{\textcolor{blue}{10.2}} & \textbf{\textcolor{blue}{1.20}} &  11.7 \\
         \midrule
        2 &  PASMNet~\cite{wang2020parallax} & \XSolidBrush & $\textit{\textbf{d}}$ & 1.36 & 8.17 & 1.74 & 13.9 & 2.10 & 15.7 & 1.90 & 14.4 \\
         3 & PASMNet~\cite{wang2020parallax} & \XSolidBrush & $\textit{\textbf{dT(nfr)}}$ & 1.82 & 11.3 & 2.04 & 16.8 & 3.15 & 23.2 &  2.29 & 18.7 \\
         4 &  ChiT~\cite{su2022chitransformer} & \XSolidBrush & $\textit{\textbf{d}}$ & 1.19 & 6.16 & 1.37 & 11.1 & {1.37} & {10.9} & 1.74 & 12.8 \\
         5 & ChiT~\cite{su2022chitransformer} & \XSolidBrush &  $\textit{\textbf{dT(nfr)}}$ & 1.27 & 8.07 & 1.69 & 14.0 & 1.95 & 12.3 & 1.95 & 14.6\\
         6 & DTD~\cite{sharma2020nighttime} & \XSolidBrush &  $\textit{\textbf{T(n)}}$ & - & - & - & - & - & - &  {1.28} &  \textbf{\textcolor{blue}{7.36}} \\
         \midrule
         7 & \textbf{RoSe} & \XSolidBrush & $\textit{\textbf{d}}$ & \textbf{0.80} & \textbf{1.77} & {1.27} & \textbf{\textcolor{blue}{10.6}} & {1.21} & {10.4} & {1.63} & {9.35}\\
         8 & \textbf{RoSe} & \XSolidBrush & $\textit{\textbf{dT(nfr)}}$ & \textbf{\textcolor{blue}{0.83}} 
    & \textbf{\textcolor{blue}{1.84}} & \textbf{0.87} & \textbf{2.04} & \textbf{0.92} & \textbf{2.16} & \textbf{1.04} & \textbf{5.48} \\
         \bottomrule
    \end{tabular}}
    \vspace{-0.2cm}
    \label{tab:ms2}
\end{table*}

\begin{figure*}[t]
    \includegraphics[width=1.0\linewidth]{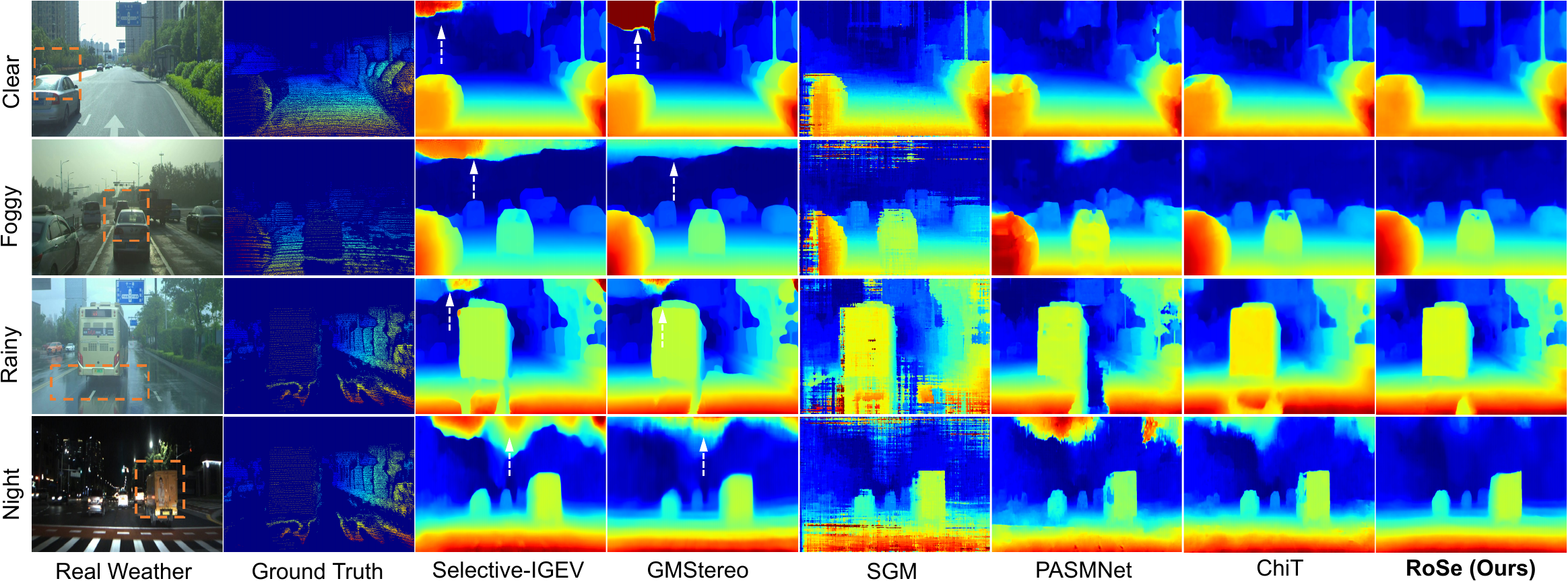}
    \vspace{-0.6cm}
    \caption{Visual comparison on DrivingStereo and MS2 validation set. We utilize the officially provided codebases to retrain the models on the Adverse-DrivingStereo and Adverse-MS2 datasets.
    Note that the night condition is from the MS2 dataset.}
    \label{sec4:visual_comparison}
    \vspace{-0.3cm}
\end{figure*}

\subsection{Implementation Details.} 
The training setup for these three datasets involves three stages: pre-training, self-supervised scene correspondence learning, and adverse weather distillation.
1) Pre-training: The network is initially pre-trained on the synthetic SceneFlow dataset~\cite{mayer2016large} in a supervised manner.
This step uses a batch size of 32 and a learning rate of $2\times10^{-4}$ for 20k iterations to provide reasonable pre-trained parameters for subsequent self-supervised learning.
Asymmetric augmentation~\cite{yang2019hierarchical} is used to prevent the model from overfitting.
We have empirically shown that having well-trained parameters is critical for effective self-supervised learning in stereo matching, as illustrated in Sec.~\ref{sec4:ablation_study}.
2) Self-supervised scene correspondence learning: The network is then fine-tuned on the Adverse-DrivingStereo, Adverse-MS2, and Adverse-KITTI training datasets using the proposed self-supervised framework and losses, and evaluate its performance on the corresponding validation weather sets.
This step employs an initial learning rate of $2\times10^{-5}$ with a batch size of 16 for 20k iterations.
Random color and illumination transformations are used for data augmentation. 
3) Adverse weather distillation: In this final step, the architecture of the student model mirrors that of the teacher model, but the weights are re-trained.
This step employs an initial learning rate of $2\times10^{-4}$ with a batch size of 16 for 20k iterations.
Beyond employing random color and illumination transformations, we utilize asymmetric augmentation to simulate failure cases in photometric consistency for data augmentation. This data augmentation strategy helps improve the model's robustness by imposing potential inconsistencies across views.
Throughout these steps, images are randomly cropped to a size of
$384\times832$. 
All training is conducted using two RTX A100 GPUs in the PyTorch platform. 
We utilize the AdamW optimizer with $\beta_{1} = 0.9$, $\beta_{2} = 0.999$, and a cosine learning rate scheduler with a warm-up phase to optimize the model.

\subsection{Comparison to SoTA Methods}
\label{sec4:sota}
To our knowledge, among self-supervised methods for challenging conditions, only DTD~\cite{vankadari2024dtd} offers open-source code for nighttime stereo depth estimation. 
Therefore, to ensure a fair comparison, we will consider only works that have publicly available code.
We compare our method with a classical method, Semi-Global Matching (SGM)~\cite{hirschmuller2007stereo}, two latest open-source self-supervised SoTA methods: PASMNet~\cite{wang2020parallax} and ChiTransformer~\cite{su2022chitransformer}, an open-source self-supervised SoTA method applied at night: DTD~\cite{vankadari2024dtd}, and two open-source supervised SoTA methods: GMStereo~\cite{xu2023unifying}, and selective-IGEV~\cite{wang2024selective}.
We utilized the officially provided codes to re-train these methods on the same training sets as our methods to ensure a fair comparison.
In addition, we also compare our method with standard self-supervised methods on the KITTI benchmark. 

\subsubsection{Robust Zero-shot Generalization}
\revise{We first conduct experiments to demonstrate that our pre-trained model on the synthetic Scene Flow dataset generalizes well to diverse real-world weather conditions in a zero-shot setting.}
As illustrated in Table~\ref{sec4:dg} and Table~\ref{sec4:dg_comparison}, our pre-trained {RoSe} almost outperforms most generalized SoTA domain methods, and the improvement is more obvious in adverse conditions. 
Unlike the domain-generalized stereo method Former-PSMNet~\cite{zhanglearning2024}, our model does not incorporate any custom-tailored modules or specifically designed losses for generalization.
Additionally, the capacity of our VFM is ViT-Base, in contrast to Former-PSMNet, which utilizes ViT-Large. This comparison highlights our model's efficiency and effectiveness even with a more modest architecture.

\begin{table}[t]
    \scriptsize
    \caption{Compared with fully-supervised stereo methods on the DrivingStereo and MS2 validation weather set. Bad 3.0 metric is adopted. Note that these two stereo methods are fine-tuned on these two training sets in a supervised manner.}
    \vspace{-0.2cm}
    \centering    
    \setlength{\tabcolsep}{1mm}{
    \begin{tabular}{c|cc|ccc|ccc}
        \toprule
        \multirow{2}{*}{\textbf{ID}}  & \multirow{2}{*}{\textbf{Method}}  & \multirow{2}{*}{\textbf{Sup.}} & \multicolumn{3}{c|}{\textbf{DrivingStereo}} &\multicolumn{3}{c}{\textbf{MS2}}\\
        & & & {\textbf{Clear}}  & {\textbf{Foggy}} & {\textbf{Rainy}} & {\textbf{Clear}} & {\textbf{Rainy}} & {\textbf{Night}} \\
         \midrule
         1 & GMStereo~\cite{xu2023unifying} & \Checkmark & \textbf{\textcolor{blue}{1.22}}  & \textbf{\textcolor{blue}{0.91}} & \textbf{\textcolor{blue}{0.99}} & \textbf{\textcolor{blue}{1.31}} & \textbf{\textcolor{blue}{1.40}} & \textbf{2.93} \\
         2 & Selective-IGEV~\cite{wang2024selective} & \Checkmark   & \textbf{1.04}  & \textbf{0.78} & \textbf{0.85} & \textbf{1.18} & \textbf{1.24} & \textbf{\textcolor{blue}{3.15}} \\
         \midrule
        3 & \textbf{RoSe (Ours)} & \XSolidBrush & {1.60} & {1.71} & {1.88} & 1.84 & 2.16 & 5.48 \\
         \bottomrule
    \end{tabular}}
    \label{tab:sup}
    \vspace{-0.4cm}
\end{table}

\begin{figure}[t]
\centering
\includegraphics[width=1.\linewidth]{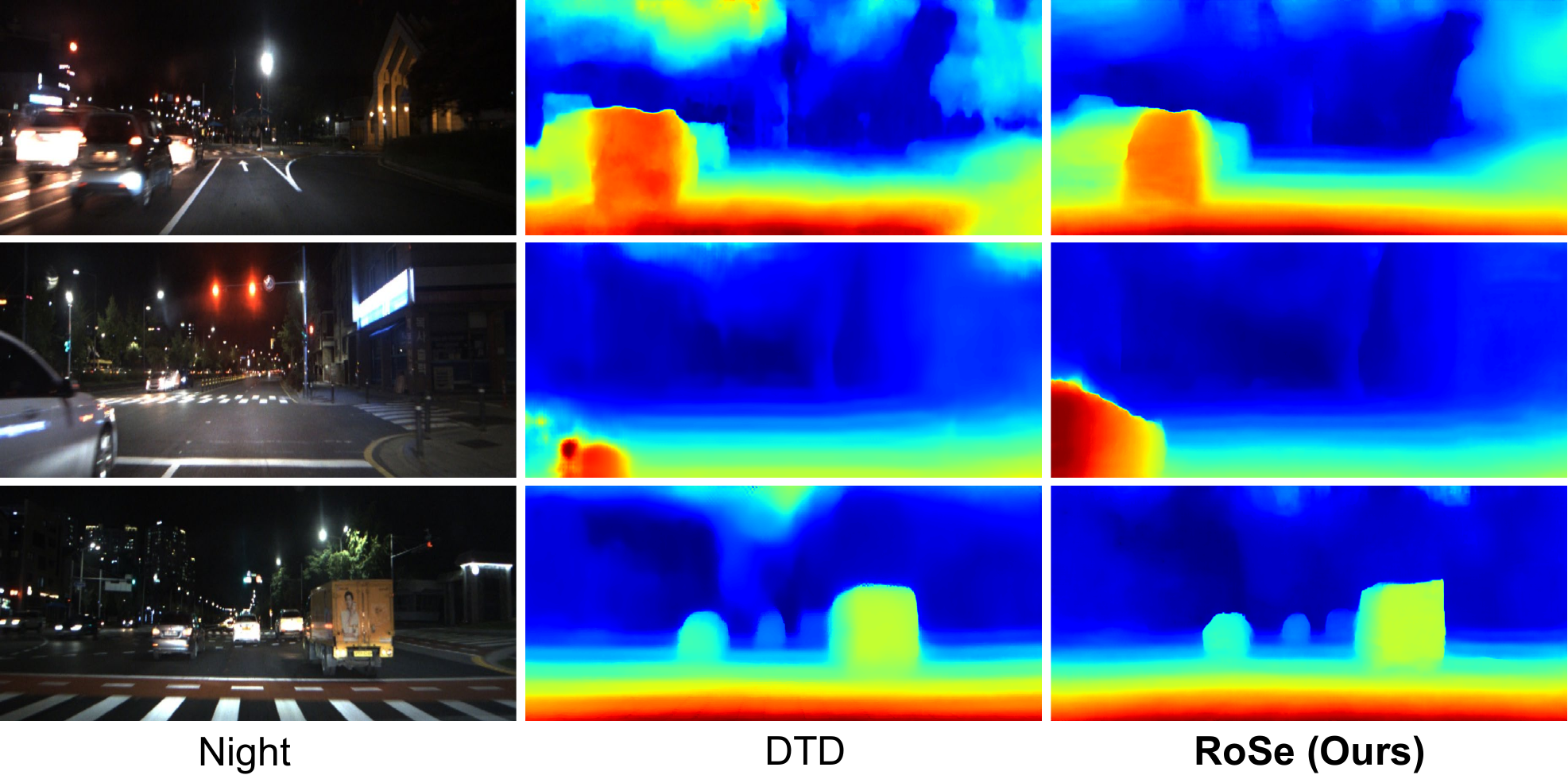}
\vspace{-0.6cm}
\caption{ The qualitative comparison of the proposed method DTD~\cite{vankadari2024dtd} and ours on MS2 night validation set.}
\label{sec4:dtd}
\vspace{-0.5cm}
\end{figure}

\begin{table*}[t]
\caption{Comparative results achieved on the KITTI 2012 \& 2015 benchmarks. The latest state-of-the-art self-supervised stereo matching method is ChiT but using KITTI Eigen Splits (22600 image pairs), denoted as $\ast$.
“-” indicates that results are not available.}
\vspace{-0.5cm}
    \begin{center}
    \scriptsize
	\setlength{\tabcolsep}{2.5mm}{
		\begin{tabular}{c|c|cccc|ccc|c}
		\toprule
		& & \multicolumn{4}{c|} {\textbf{KITTI 2012}}  & \multicolumn{3}{c}{\textbf{KITTI 2015}}  \\
		\multirow{1}{*}[5pt]{Method} & \multirow{1}{*}[5pt]{Sup.} & \textbf{Out-Noc (\%)} & \textbf{Out-All (\%)} & \textbf{Avg-Noc (px)} & \textbf{Avg-All (px)} & \textbf{D1-bg (\%)} & \textbf{D1-fg (\%)} & \textbf{D1-All (\%)} & \textbf{Time (s)}\\
		\midrule
Zhou et al.~\cite{zhou2017unsupervised} & - & - & - & - & - & - & 9.91 & 8.61 & - \\
OASM~\cite{li2018occlusion} & \XSolidBrush & 6.39 & 8.60 & 1.3 & 2.0 & 6.89 & 19.42 & 8.98 & 0.73 \\
PASMnet\_192~\cite{wang2020parallax} & \XSolidBrush & 7.14 & 8.57 & 1.3 & 1.5 & 5.41 & 16.36 & 7.23 & 0.5 \\
Flow2Stereo~\cite{liu2020flow2stereo} & \XSolidBrush & 4.58 & 5.11 & 1.0 & 1.1 & 5.01 & 14.62 & 6.61 & 0.05 \\
DispSegNet~\cite{zhang2019dispsegnet} & \XSolidBrush & 4.68 & 5.66 & 0.9 & 1.0 & 4.20 & 16.67 & 6.33 & 0.9 \\
    Reversing-PSMNet~\cite{aleotti2020reversing} & \XSolidBrush & - & - & - & - & 3.13 & 8.70 & 4.06 &  0.41 \\
ChiT$^{\ast}$~\cite{su2022chitransformer} & \XSolidBrush & - & -& - & - &  \textbf{2.50} & \textbf{\textcolor{blue}{5.49}}  & \textbf{\textcolor{blue}{3.03}} & 0.32 \\
\midrule
    \textbf{RoSe (Ours)} & \XSolidBrush & \textbf{2.55} & \textbf{3.17} & \textbf{0.7} & \textbf{0.8} & \textbf{\textcolor{blue}{2.65}} &  \textbf{4.36} &  \textbf{2.94} &\textbf{0.17} \\
\bottomrule
\end{tabular}}
\end{center}
\vspace{-0.4cm}
\label{sec4:table_kitti}
\end{table*}

\subsubsection{Adverse Weather}
\revise{In this section, we evaluate two challenging datasets with diverse real-world adverse weather conditions. The DrivingStereo dataset includes clear, rainy, and foggy conditions but lacks nighttime scenarios. Conversely, the MS2 dataset includes nighttime conditions but lacks foggy weather scenarios. To address these gaps, we use the CycleGAN-Turbo model~\cite{parmar2024one} to generate synthetic weather sets for their respective missing weather conditions, ensuring comprehensive evaluation across all relevant scenarios.}

\textbf{DrivingStereo.}
In Table~\ref{tab:dr}, we report results for DrivingStereo across various weather settings. 
It is clear that our method {RoSe} achieves remarkable performance across various weather and road conditions.
A surprising result is that self-supervised methods trained only on clear data (ID = 2 \& 4) surpass the performance of those trained on mixed data (ID = 3 \& 5) by a large margin. This reveals that the previous SoTA self-supervised methods based on the photometric consistency loss are susceptible to adverse weather, resulting in training collapse.
Moreover, compared with the classic method SGM (ID = 1), these self-supervised methods are more sensitive in rainy and night scenes, which exhibit poor generalization ability.
In contrast, benefiting from the proposed robust self-supervised framework, our method can be trained on these adverse conditions (ID = 8) and significantly outperforms all SoTA self-supervised methods.
Moreover, as illustrated in Fig.~\ref{sec4:visual_comparison}, our method makes robust estimations in challenging regions denoted as orange dashed boxes.
Overall, qualitative and quantitative results demonstrate the robustness of our model to a variety of adverse conditions.

\textbf{MS2.}
To further validate the generalisability of our method,
we evaluate these models using the official weather test set of the MS2 dataset. 
As shown in Table~\ref{tab:ms2}, similarly, when training self-supervised methods with the adverse
weather data (ID = 3 \& 5), we obtain a disappointing performance, indicating a lack of ability to handle adverse weather conditions.
Remarkably,
our method significantly improves performance in all adverse weather conditions without requiring specialized architectures (ID = 8).
Overall, our method achieves the best performance among SoTA self-supervised stereo methods.

\begin{figure*}[t]
    \includegraphics[width=1.0\linewidth]{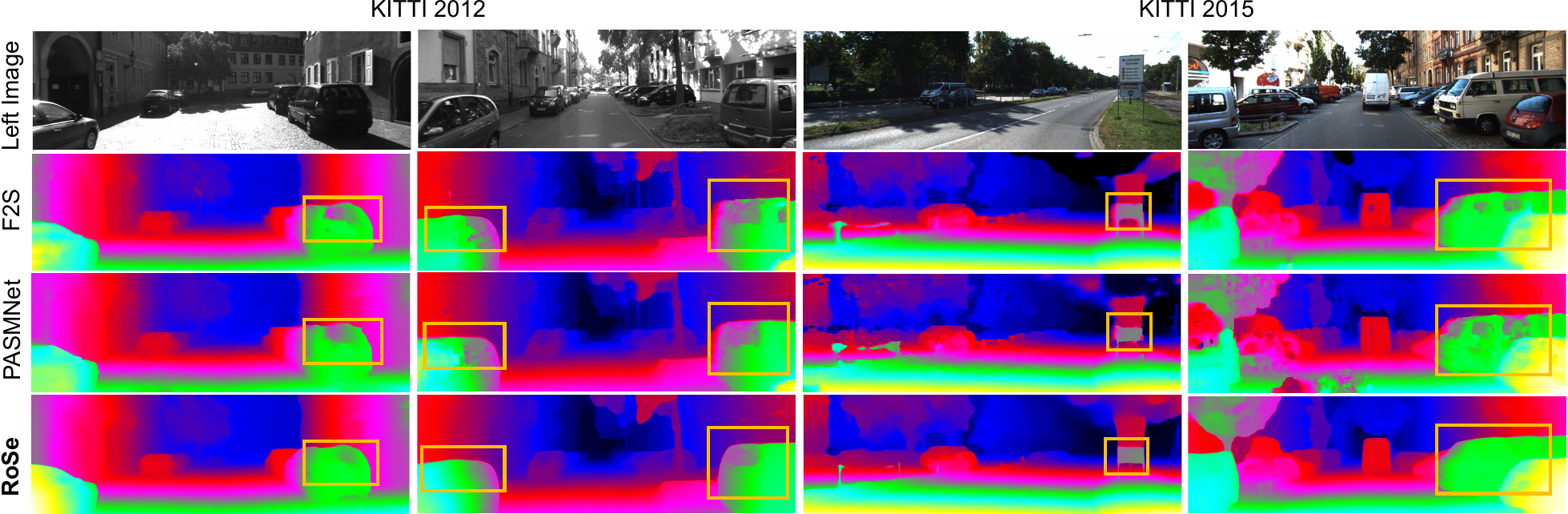}
    \vspace{-0.6cm}
    \caption{Visualization results achieved by our method and other self-supervised methods from KITTI 2015 \& KITTI 2012.}
    \label{sec4:vis_kitti}
    \vspace{-0.4cm}
\end{figure*}

\textbf{Comparison to Supervised Methods.}
To further verify the superiority of the proposed robust self-supervised framework, we also compare our method with supervised SoTA stereo networks (Selective-IGEV and GMStereo) on validation weather sets. 
As shown in Table~\ref{tab:sup}, our method is slightly worse than theirs.
Interestingly, from Fig.~\ref{sec4:visual_comparison}, it seems that the qualitative results of GMStereo and Selective-IGEV (3rd \& 4th column) produce artifacts in areas without GT, such as the sky, water spots, etc. By contrast, our method rarely has this problem and generates smoother disparity maps (the last column).

\begin{figure*}[t]
\centering
\includegraphics[width=1.\linewidth]{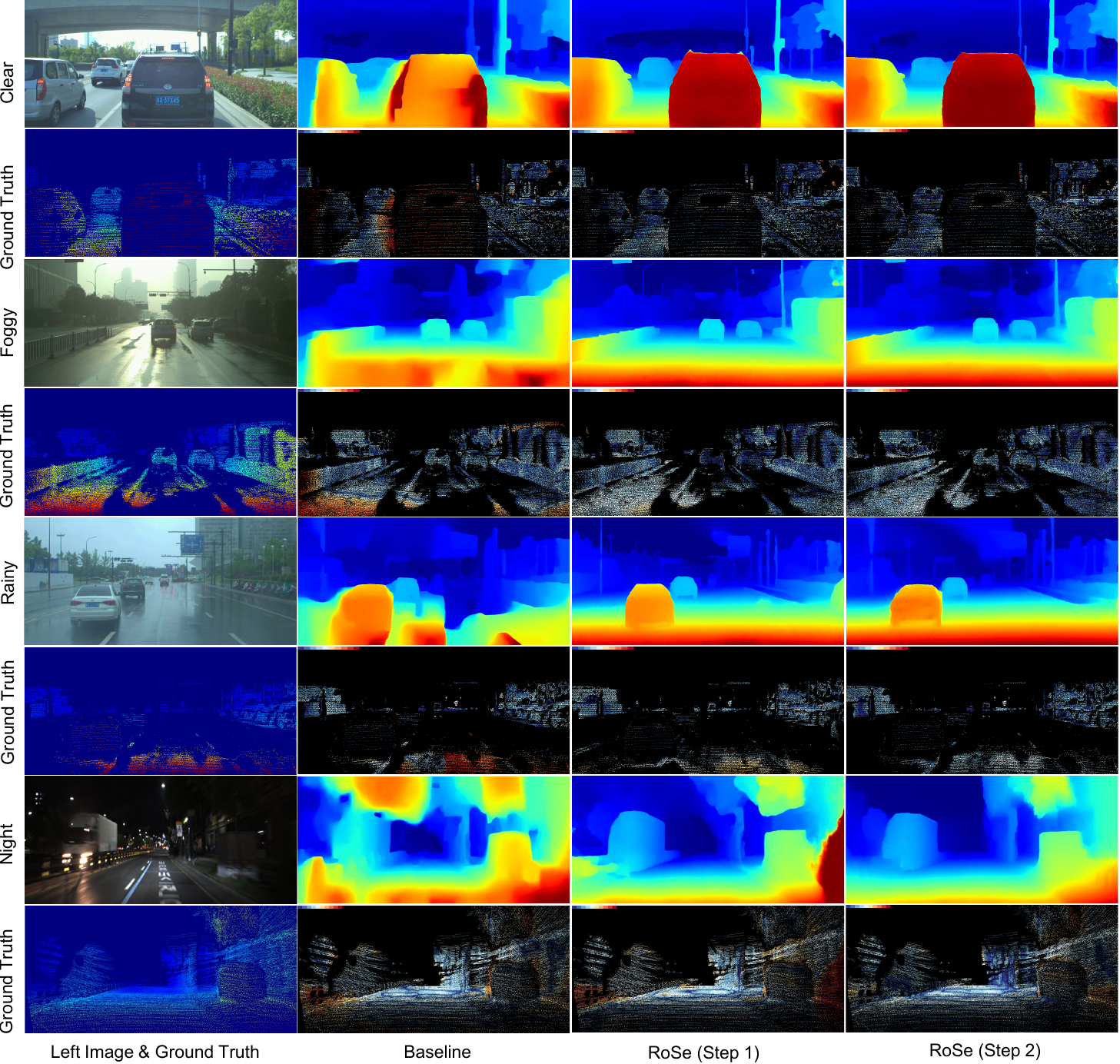}
\vspace{-0.4cm}
\caption{ Visual comparison of RoSe on the weather validation set. Baseline denotes that RAFTStereo~\cite{Lipson2021RAFTStereoMR} was trained using the vanilla photometric consistency and disparity smooth losses. Step 1 indicates the self-supervised scene corresponding learning step. Step 2 denotes the adverse weather distillation step. The predicted disparity maps and their corresponding error maps are displayed in columns 2, 3, and 4, respectively.}
\label{sec4:ablation_comparison}
\vspace{-0.2cm}
\end{figure*}

\subsubsection{Visual Comparison with DTD}
\label{DTD}
DTD~\cite{vankadari2024dtd}
is an open-source algorithm designed to focus on nighttime
conditions in self-supervised stereo matching.
To make a fair comparison, we follow the default settings and use the officially provided code to retrain the model on the Adverse-DrivingStereo night training set and MS2 night training set.
We provide visual example comparisons in Fig.~\ref{sec4:dtd}. As we can see, our method recovers more details in challenging regions.

\subsubsection{Results on KITTI benchmark}
We also conducted evaluations on the KITTI benchmarks. Specifically, we train our model exclusively on the Adverse-KITTI datasets and evaluate the KITTI test set.
Table~\ref{sec4:table_kitti} shows that our method outperforms previous standard self-supervised methods by notable margins. It is important to note that our method is adept at handling challenging scenarios while exhibiting superior performance in clear scenarios (KITTI dataset), showcasing its effectiveness and universality.
The qualitative results, as shown in Fig.~\ref{sec4:vis_kitti}, further illustrate our method's excellence in recovering difficult regions. 
Additionally, while the latest state-of-the-art method, ChiT~\cite{su2022chitransformer}, utilizes KITTI eigen splits with 22600 image pairs for training, our method, following the regular training setup~\cite{wang2020parallax,li2018occlusion}, outperforms ChiT using only 394 image pairs (i.e., 394 image pairs with different weather settings) from a mixture of KITTI 12 and 15 datasets.
This highlights the superior efficiency and effectiveness of our approach in achieving high-quality stereo matching results with significantly fewer training samples.

\begin{table*}[t]
\centering
\footnotesize
\caption{Ablation study on the proposed components. We train the adverse-Drivingstereo dataset and evaluate the model on the Drivingstereo and MS2 weather validation sets.
We adopt RAFTStereo~\cite{Lipson2021RAFTStereoMR} as the baseline.
FPN denotes the Feature Pyramid Network, and DAMv2~\cite{depth_anything_v2} is adopted as VFM.
AFEM denotes the proposed anti-adverse feature enhancement module.
\textcolor{orange}{Orange} color in the table represents the results of the baseline, while \textcolor{pink}{pink} color represents the results of the final model.}
\vspace{-0.2cm}
\setlength{\tabcolsep}{1mm}{
    \begin{tabular}{l|c|cccc|c|cc|cc|cc|cc}
    \toprule
    \multirow{2}{*}[-1pt]{\textbf{Model}} & {\textbf{Feature}} & \multicolumn{4}{c|}{\textbf{Step 1}} & \textbf{Step 2} & \multicolumn{2}{c|}{\textbf{Clear}} & \multicolumn{2}{c|}{\textbf{Foggy}} & \multicolumn{2}{c|}{\textbf{Rainy}} & \multicolumn{2}{c}{\textbf{Night (MS2)}} \\
    & \textbf{Extractor} & {$\mathcal{L}_{photo}$} & {$\mathcal{L}_{s}$} & {$\mathcal{L}_{fc}$} & {$\mathcal{L}_{dc}$}  & {$\mathcal{L}_{kd}$} & \textbf{EPE}$\downarrow$  & \textbf{Bad 3.0}$\downarrow$  & \textbf{EPE} $\downarrow$  & \textbf{Bad 3.0} $\downarrow$  & \textbf{EPE}$\downarrow$  & \textbf{Bad 3.0}$\downarrow$   & \textbf{EPE} $\downarrow$  & \textbf{Bad 3.0} $\downarrow$  \\
    \midrule
    Baseline~\cite{Lipson2021RAFTStereoMR} & FPN & \Checkmark & \Checkmark &  & &  &  \cellcolor{orange} 1.74 & \cellcolor{orange} 7.87 & \cellcolor{orange} 2.25  & \cellcolor{orange} 9.92 & \cellcolor{orange} 2.46 & \cellcolor{orange} 10.8 & \cellcolor{orange} 3.19 & \cellcolor{orange} 19.6 \\
    \midrule
    \multirow{6}{*}{\textbf{RoSe}} & VFM & \Checkmark & \Checkmark & & & & 1.89 & 8.31 & 1.96 & 9.17 & 2.09 & 9.52 & 2.68 & 17.1 \\
    & FPN + VFM & \Checkmark & \Checkmark &  & & & 1.54 & 5.49 & 1.75 & 7.12 & 1.84 & 7.40 & 2.25 & 15.3 \\
    & FPN + VFM & \Checkmark & \Checkmark & & \Checkmark & & 1.26 & 3.73 & 1.34 & 3.89 & 1.41 & 4.03 & 2.11 & 14.2 \\
    & FPN + VFM & \Checkmark & \Checkmark & \Checkmark & \Checkmark & & 1.01 & 2.54 & 1.09 & 2.76 & 1.13 & 2.87 & 1.94 & 13.5 \\
    & FPN + VFM + AFEM & \Checkmark & \Checkmark & \Checkmark & \Checkmark & & \textbf{\textcolor{blue}{0.91}} & \textbf{\textcolor{blue}{2.09}} & \textbf{\textcolor{blue}{0.98}} & \textbf{\textcolor{blue}{2.25}} & \textbf{\textcolor{blue}{1.01}} & \textbf{\textcolor{blue}{2.30}} & \textbf{\textcolor{blue}{1.69}} & \textbf{\textcolor{blue}{11.4}} \\
    & FPN + VFM + AFEM & \Checkmark & \Checkmark & \Checkmark & \Checkmark & \Checkmark & \cellcolor{pink} $\textbf{0.78}$ & \cellcolor{pink} $\textbf{1.60}$ & \cellcolor{pink} $\textbf{0.85}$ & \cellcolor{pink} $\textbf{1.71}$ & \cellcolor{pink} $\textbf{0.88}$ & \cellcolor{pink} $\textbf{1.88}$ & \cellcolor{pink} $\textbf{1.57}$ & \cellcolor{pink} $\textbf{10.8}$ \\
    \bottomrule
    \end{tabular}}
    \vspace{-0.3cm}
\label{sec4:ablation}
\end{table*}

\subsubsection{Model Efficiency}
The model efficiency is evaluated based on average latency and memory consumption, as is illustrated in Table~\ref{tab:dr} (the last two columns). 
To ensure a fair comparison, we evaluate the efficiency metrics of different methods using the same GPU device.
For the Drivingstereo dataset, metrics are measured on image pairs of size $400 \times 881$ on a single A100 GPU, excluding the first inference for steady-state measurements. 
For the KITTI dataset, metrics are measured on image pairs of size $384 \times 1248$ on a single A100 GPU. 
As we can see, our method shows reasonable memory consumption and fast inference time.

\begin{table}[t]
    \scriptsize
    \caption{Zero-shot performance with various VFM backbones of different capacities on the SceneFlow dataset is evaluated using the Bad 3.0 metric. Note that these model variants are trained on the synthetic dataset in a supervised manner.}
    \vspace{-0.1cm}
    \centering    
    \setlength{\tabcolsep}{0.2mm}{
    \begin{tabular}{cc|ccc|c|cc}
        \toprule
     \textbf{VFM} & {\textbf{Capacity}}  & {\textbf{Clear} $\downarrow$ }  & {\textbf{Foggy} $\downarrow$ } & {\textbf{Rainy} $\downarrow$ } & {\textbf{Night (MS2)} $\downarrow$ } & Memory & Time\\
         \midrule
        SAM~\cite{kirillov2023segment} & Small & 4.17 & 5.22 & 4.91 & 21.2 & 2.4 GB & 0.13 s \\
        SAM~\cite{kirillov2023segment} & Base & 3.81 & 4.96 & 4.69 & 20.6 & 2.8 GB & 0.15 s \\
        SAM~\cite{kirillov2023segment} & Large & \textbf{3.47} & \textbf{4.75} &  \textbf{4.34} &  \textbf{20.1} & 3.7 GB & 0.20 s\\
         \midrule
          DAMV2~\cite{depth_anything_v2} & Small & {3.97} & {5.16} & {5.07} & 21.3 & 2.3 GB & 0.11 s\\
          DAMV2~\cite{depth_anything_v2} & Base & {3.63} & {4.85} & {4.75} & 20.8 & 2.7 GB & 0.13 s\\
          DAMV2~\cite{depth_anything_v2} & Large & \textbf{3.22}  & \textbf{4.69} & \textbf{4.59} & \textbf{20.3} & 3.5 GB & 0.18 s\\
         \bottomrule
    \end{tabular}}
    \vspace{-0.3cm}
    \label{tab:vfm}
\end{table}

\subsection{Ablation Study}
\label{sec4:ablation_study}
We conduct ablation studies by training these model variants on the Adverse-DrivingStereo training dataset. 
We then verify the effectiveness of each component on its real weather validation set (clear, foggy, and rainy) and the MS2 night test set.
The baseline results are shown in Table~\ref{sec4:ablation} (Row 1).

\textbf{Robust Feature Extractor (Row 2 \& 3):}
Compared with baseline (Row 1), VFM (Row 2) as the feature extractor shows robustness under challenging weather conditions (rainy and night).
By combining with FPN as the feature extractor, the model variant (Row 3) further achieves improved performance, indicating the effectiveness of our design.

\textbf{Scene Correspondence Learning (Row 4 \& 5):}
Building on the robust feature extractor, we introduce the feature consistency loss $\mathcal{L}_{fc}$ and disparity correspondence loss $\mathcal{L}_{dc}$ to manage adverse weather data. Implementing these proposed losses results in a significant reduction in both EPE and Bad 3.0 metrics. This fully demonstrates the effectiveness of the proposed self-supervised learning mechanism in enhancing the network's robustness against adverse conditions.

\textbf{Anti-adverse Feature Enhancement Module (Row 6):}
Compared to the model variant in Row 5, the proposed Adverse Feature Extraction Module (AFEM) in Row 6 is designed to generate degradation-invariant features under challenging weather conditions such as foggy, rainy, and night environments. Supervised by these two consistency losses, this approach treats adverse weather degradations as diverse image styles and extracts consistent feature representations, aligning with those obtained from clear images by the original feature extractor. 
By integrating AFEM into the feature extractor, subsequent model variants demonstrate enhanced robustness, indicating the efficacy of our design.

\begin{figure}[t]
\centering
\includegraphics[width=1.\linewidth]{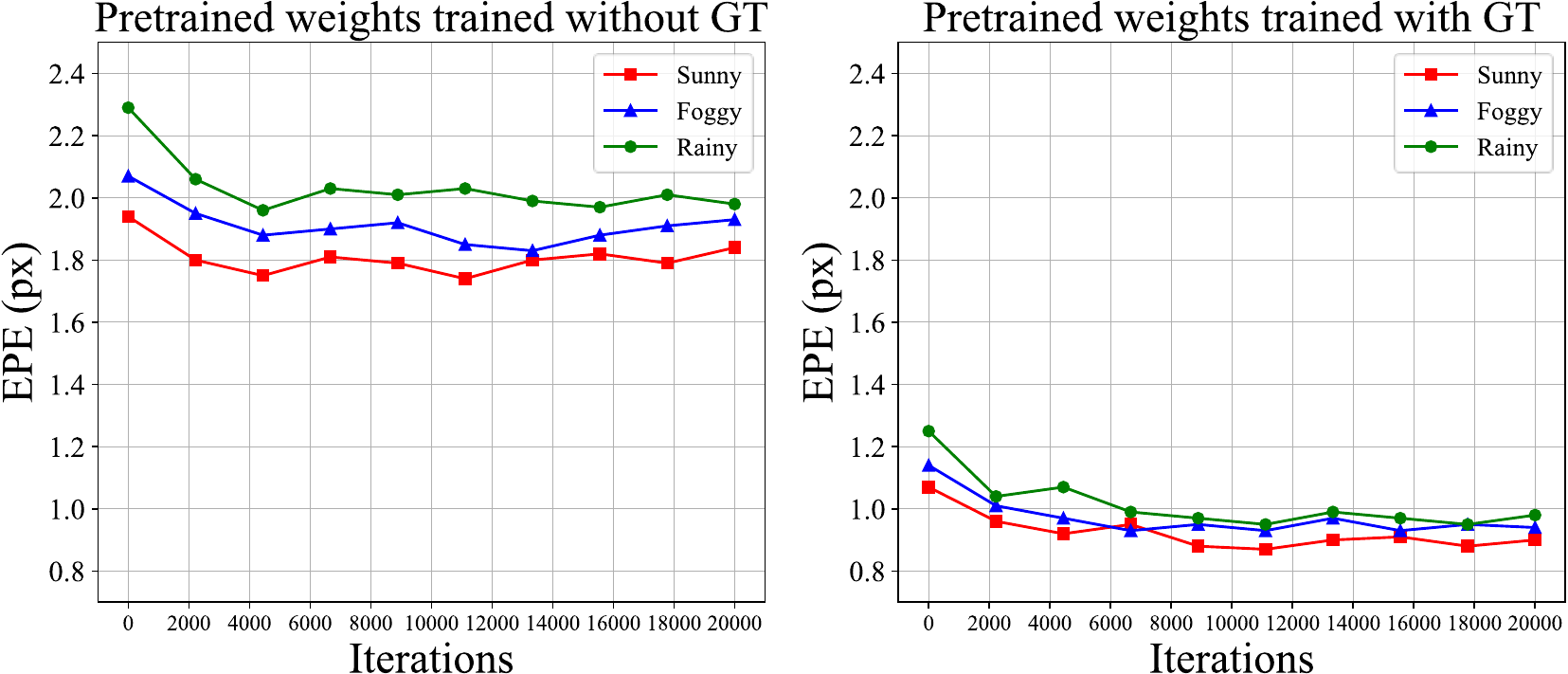}
\vspace{-0.6cm}
\caption{Comparison with different pre-trained weights on DrivingStereo weather validation datasets. EPE results are reported.}
\label{sec4:epe_curve}
\vspace{-0.3cm}
\end{figure}
\textbf{Adverse Weather Distillation (Row 7):}
By implementing the knowledge distillation strategy, we further enhance performance. The robust self-supervised model from the initial step provides high-quality pseudo labels that serve as Ground Truth (GT) for the training model, leading to significant performance improvements across mixed weather data. This demonstrates the effectiveness of our self-supervised framework.
These results emphasize the model's robustness and adaptability, highlighting its capability to maintain high performance across different challenging scenarios.

\begin{figure}[t]
\centering
\includegraphics[width=1.0\linewidth]{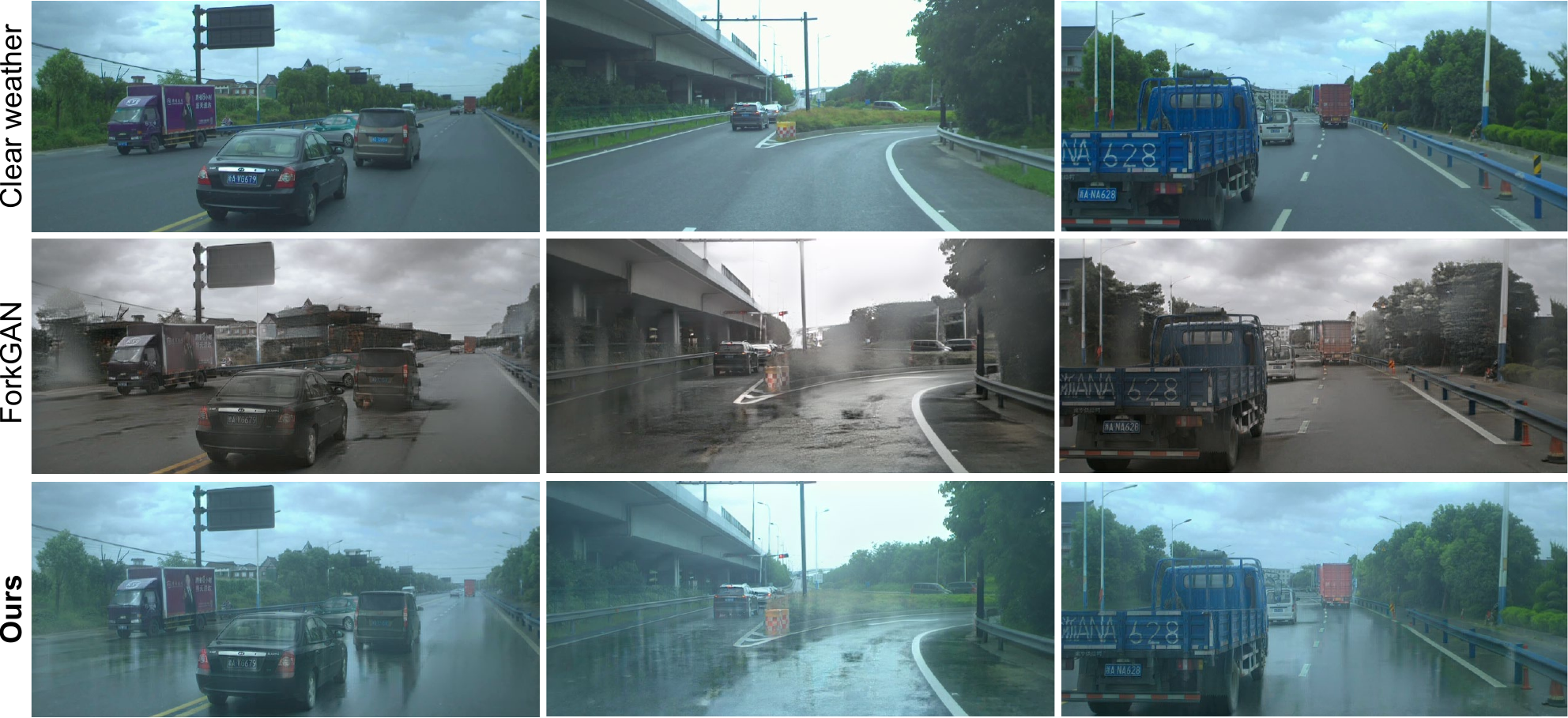}
\vspace{-0.5cm}
\caption{
Visual comparison of adverse samples (rainy) between ForkGAN~\cite{zheng2020forkgan} and the diffusion translation model~\cite{parmar2024one}.}
\label{sec4:gan_comparison}
\end{figure}

\textbf{The Effectiveness of Different VFMs.}
\label{vfm}
From Table~\ref{tab:vfm}, we make the key observation:
The larger model capacity contributes to the better robustness of the model.
Intuitively, a larger model capacity implies stronger representation ability and contains more robust priors, which are critical for the subsequent cost aggregation step.
Considering the trade-off between accuracy and computational cost, we adopt DepthAnythingV2~\cite{depth_anything_v2} (ViT-Base) as the feature backbone.

\textbf{The Impact of Pre-trained Weights.}
Fine-tuning pre-trained models to the target domain is critical for stereo matching. Previous self-supervised stereo methods~\cite{wang2020parallax, li2018occlusion} are typically pre-trained on SceneFlow in a self-supervised manner to provide pre-trained weights.
We argue that the pre-trained model trained using GT on the synthetic dataset can be relatively robust. 
Because the strong supervision from GT can encourage the model to infer reasonable disparity when faced with ill-posed regions.
To validate the assumption, we first train our RoSe on SceneFlow in a self-supervised manner. 
Then, we finetune the pre-trained weights on the adverse-DrivingStereo set as shown in Fig.~\ref{sec4:epe_curve} in the self-supervised manner (Step 1).
Remarkably, the performance of the fine-tuned model with pre-trained weights obtained in a supervised manner is much better than that of the fine-tuned model with pre-trained weights obtained in a self-supervised manner.

\begin{figure}[t]
\centering
\includegraphics[width=1.0\linewidth]{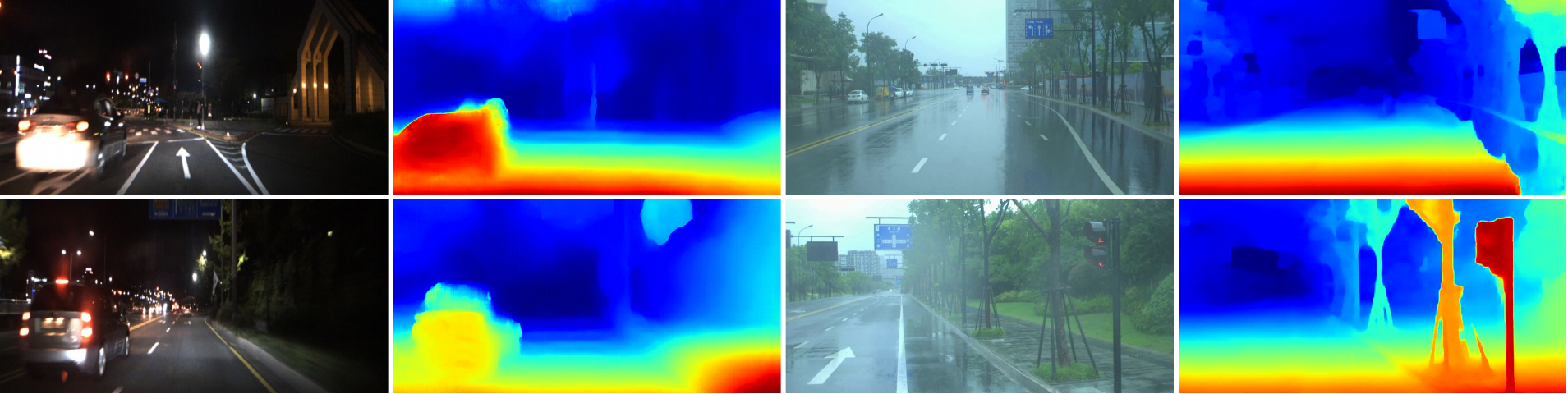}
\vspace{-0.5cm}
\caption{
Failure cases on severe image degradation scenes.}
\label{sec:failure}
\vspace{-0.2cm}
\end{figure}

\textbf{Comparison with Prior Image Translation Models.}
\textcolor{black}{We further visualize the adverse pairs (rainy pairs) generated by ForkGAN~\cite{zheng2020forkgan} and the adopted diffusion translation model~\cite{parmar2024one} we trained.
From Fig.~\ref{sec4:gan_comparison}, we observe that the adverse pairs generated using the diffusion translation model are more natural than those generated using ForkGAN~\cite{zheng2020forkgan}.}

\textbf{Robustness Against Translations.}
To assess the impact of the quality of image translation, we also use ForkGAN~\cite{zheng2020forkgan} to generate rainy pairs on the selected DrivingStereo training dataset, as shown in Table~\ref{tab:gan}.
While the selected image translation model does not give perfect outputs with full stereo consistency, it translates better than ForkGAN. 
We argue that their imperfections can enhance our model's robustness by making its disparity estimation more challenging, as the image translations serve as data augmentation strategies (e.g., asymmetric occlusion).
From Table~\ref{tab:gan}, our {RoSe} performs similarly to which image translation model is used, even when $10\%$ of the pixels of the inputs are randomly masked.

\begin{table}[t]
    \centering
    \caption{Robustness of RoSe against translations from different translation models on DrivingStereo weather set. $*$ means 10\% pixels of the rainy image pair are randomly masked~\cite{rao2023masked}.}
    \vspace{-0.2cm}
    \scriptsize
    \setlength{\tabcolsep}{2mm}{
    \begin{tabular}{c|cc|cc}
        \toprule
        \multirow{2}{*}[-1pt]{\textbf{Method}} & \multicolumn{2}{c|}{\textbf{Clear}} &  \multicolumn{2}{c}{\textbf{Rainy}} \\
        & \textbf{EPE} $\downarrow$ & \textbf{Bad 3.0} $\downarrow$ & \textbf{EPE} $\downarrow$ & \textbf{Bad 3.0} $\downarrow$ \\
        \midrule
        Ours w/ForkGAN & 0.88 & 2.07 & \textbf{\textcolor{blue}{0.90}} & \textbf{\textcolor{blue}{2.01}} \\
        Ours w/CycleGAN-Turbo& \textbf{0.84} & \textbf{1.99} & \textbf{0.84} & \textbf{1.63} \\
        Ours w/CycleGAN-Turbo$^{*}$ & \textbf{\textcolor{blue}{0.87}} & \textbf{\textcolor{blue}{2.01}} & 0.94 & 2.15 \\
        \bottomrule
    \end{tabular}}
    \label{tab:gan}
    \vspace{-0.2cm}
\end{table}

\textbf{Limitations.}
\revise{Specifically, we identify two main limitations of RoSe:
1) RoSe relies heavily on the realism and consistency of synthetic adverse weather pairs generated by image-to-image translation models. Inaccuracies or inconsistencies in these translations, such as geometric distortions or semantic mismatches, may introduce noise into the supervisory signals, potentially degrading overall performance.
2) Under severe image degradation, RoSe may produce unreliable disparity estimates (see Fig.~\ref{sec:failure}), which poses a risk in safety-critical applications like autonomous driving~\cite{Wang2025InterventionalRC}, where accurate depth perception is crucial for navigation and obstacle avoidance.}

\section{Conclusion}
\revise{In this paper, we address the performance degradation of existing self-supervised stereo matching methods under adverse weather conditions. We propose RoSe, a simple and effective solution that enables a stereo model to robustly estimate disparity by effectively addressing the weaknesses that typically hinder performance in such scenarios. Extensive experiments validate the robustness of our method, demonstrating its superiority over existing state-of-the-art solutions. Crucially, this enhanced robustness is fundamental to the safety and reliability of autonomous systems, ensuring they can operate dependably regardless of weather conditions.}


\bibliographystyle{ieeetr}
\bibliography{mybib}

\begin{IEEEbiography}[{\includegraphics[width=1in,height=1.25in,clip,keepaspectratio]{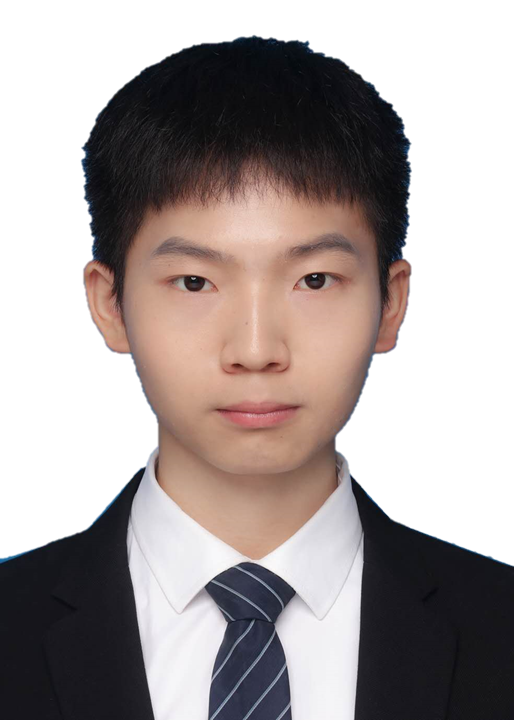}}] {Yun Wang}  received the B.E. degree from China University of Geosciences (CUG) in 2020 and the M.E.
degree from Sun Yat-sen University (SYSU), China, in 2023.
He is currently pursuing the Ph.D. degree with the Department of Computer Science, City University of Hong Kong, Hong Kong SAR. His current research interests include 3D perception and multi-model learning.
\end{IEEEbiography}

\begin{IEEEbiography}[{\includegraphics[width=1in,height=1.25in,clip,keepaspectratio]{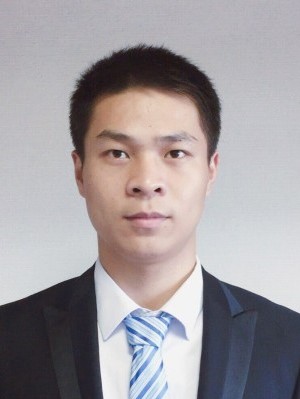}}]{Junjie Hu} (Member, IEEE) received the MS and PhD degrees from the Graduate School of Information Science, Tohoku University, Sendai, Japan, in 2017 and 2020, respectively. He is currently a research scientist with the Shenzhen Institute of Artificial Intelligence and Robotics for Society and an adjunct assistant professor at the School of Science and Engineering, the Chinese University of Hong Kong, Shenzhen. His research interests include machine learning, computer vision, and robotics. He has published more than 30 research papers in top-tier international journals and conference proceedings in AI and robotics, such as TPAMI, TRO, ICCV, IJCAI, RAL, ICRA, and IROS. He serves as a reviewer of IJCV, TIP, TNNLS, CVPR, ECCV, ICRA, IROS, RAL, IEEE Transactions on Mechatronics, Journal of Field Robotics, etc.
\end{IEEEbiography}

\begin{IEEEbiography}
[{\includegraphics[width=1in,height=1.25in,clip,keepaspectratio]{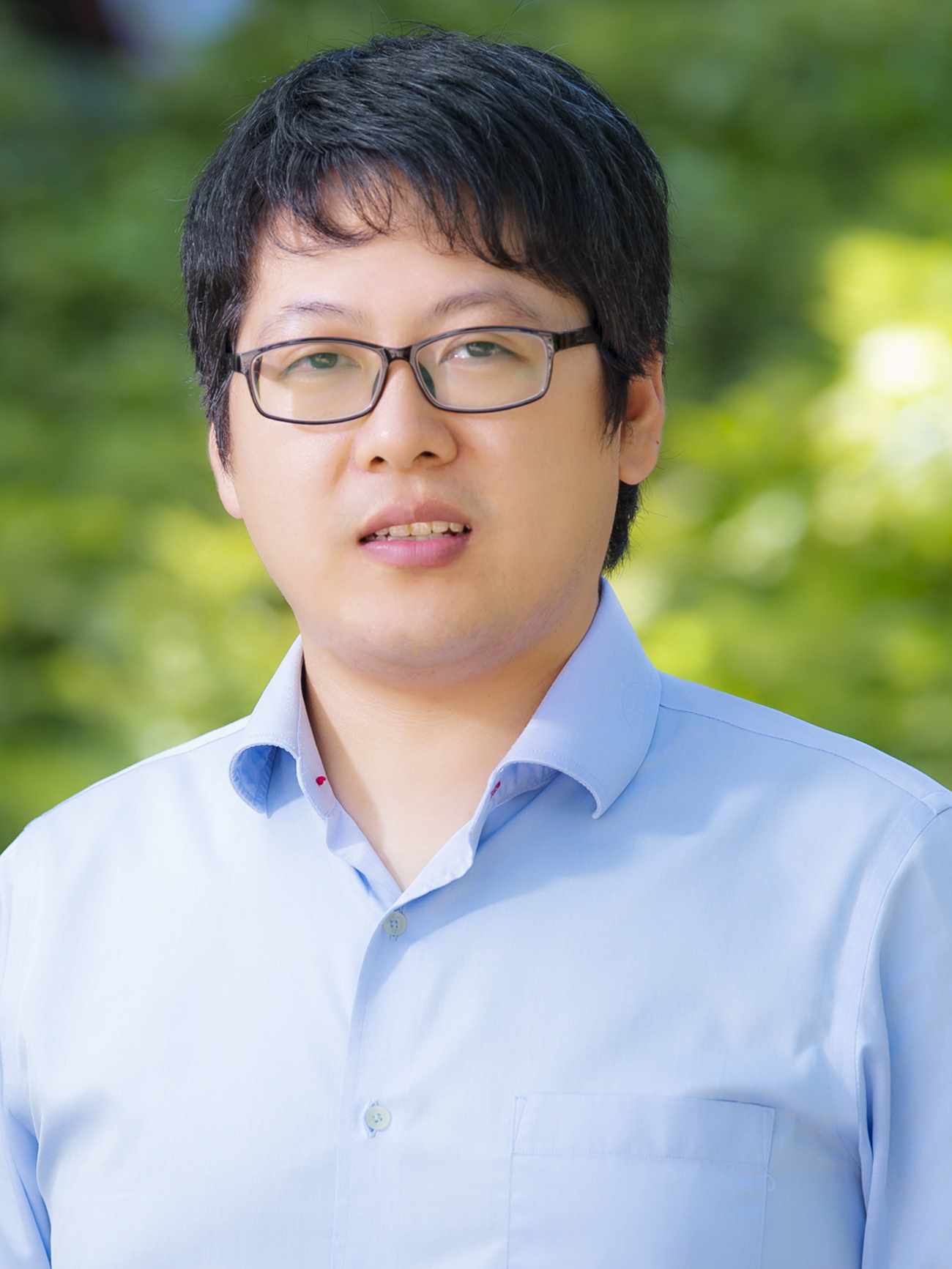}}] {Junhui Hou} (Senior Member, IEEE) is an Associate Professor with the
Department of Computer Science, City University of Hong Kong. He holds
a B.Eng. degree in information engineering (Talented Students Program)
from the South China University of Technology, Guangzhou, China (2009),
an M.Eng. degree in signal and information processing from Northwestern
Polytechnical University, Xi’an, China (2012), and a Ph.D. degree from
the School of Electrical and Electronic Engineering, Nanyang Technological
University, Singapore (2016). His research interests are multi-dimensional
visual computing.
Dr. Hou received the Early Career Award (3/381) from the Hong Kong
Research Grants Council in 2018 and the NSFC Excellent Young Scientists
Fund in 2024. He has served or is serving as an Associate Editor for IEEE
Transactions on Visualization and Computer Graphics, IEEE Transactions on
Image Processing, IEEE Transactions on Multimedia, and IEEE Transactions
on Circuits and Systems for Video Technology.
\end{IEEEbiography}

\begin{IEEEbiography}[{\includegraphics[width=1in,height=1.25in,clip,keepaspectratio]{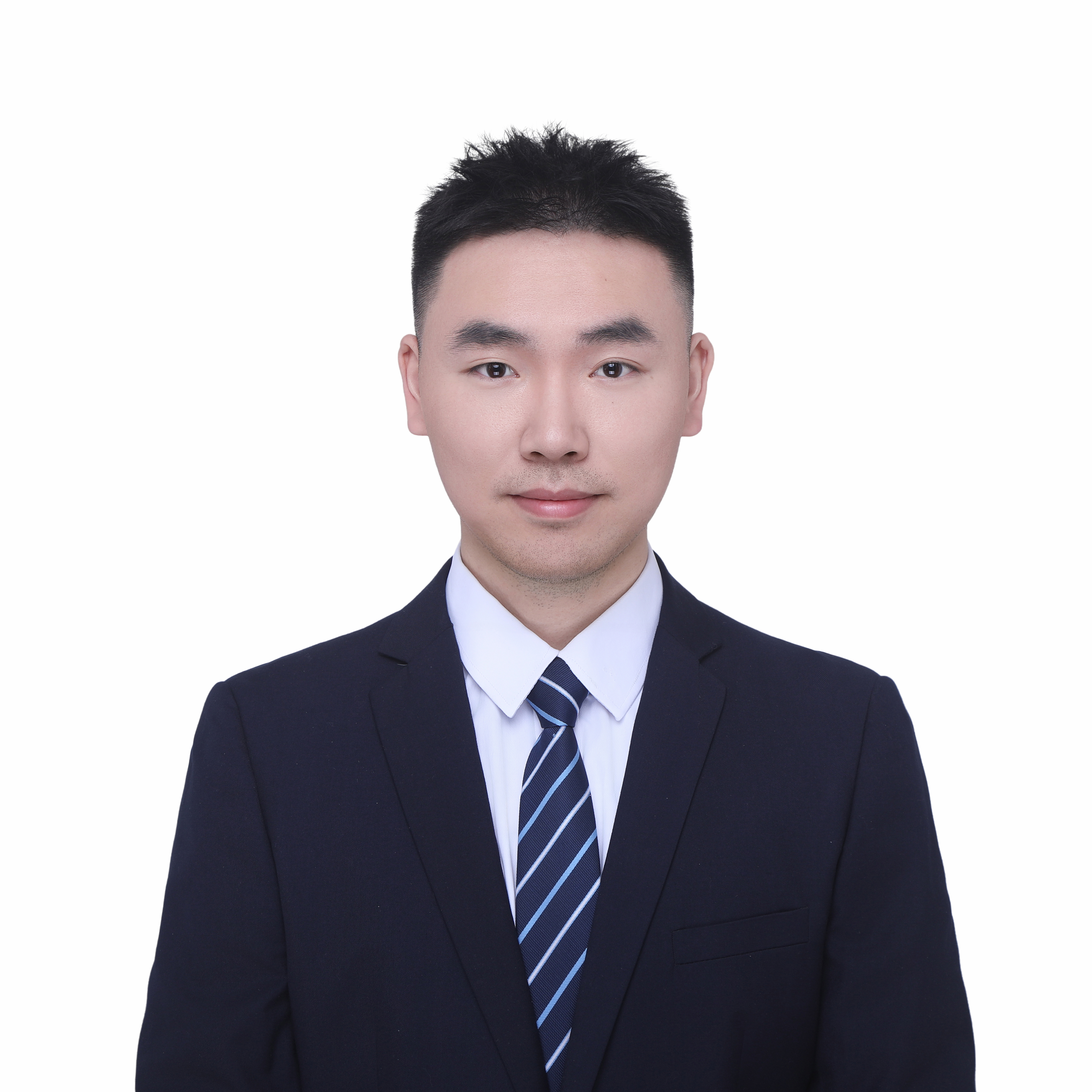}}]
{Chenghao Zhang }received the B.S. degree in software engineering from Sun Yat-Sen University, Guangzhou, China, in 2018, and the Ph.D. degree in pattern recognition and intelligent system from Institute of Automation, Chinese Academy of Sciences, Beijing, China, in 2023. He is currently a senior algorithm engineer at Alibaba Cloud. His research interests include 3D perception and vision language learning.
\end{IEEEbiography}

\begin{IEEEbiography}[{\includegraphics[width=1in,height=1.25in,clip,keepaspectratio]{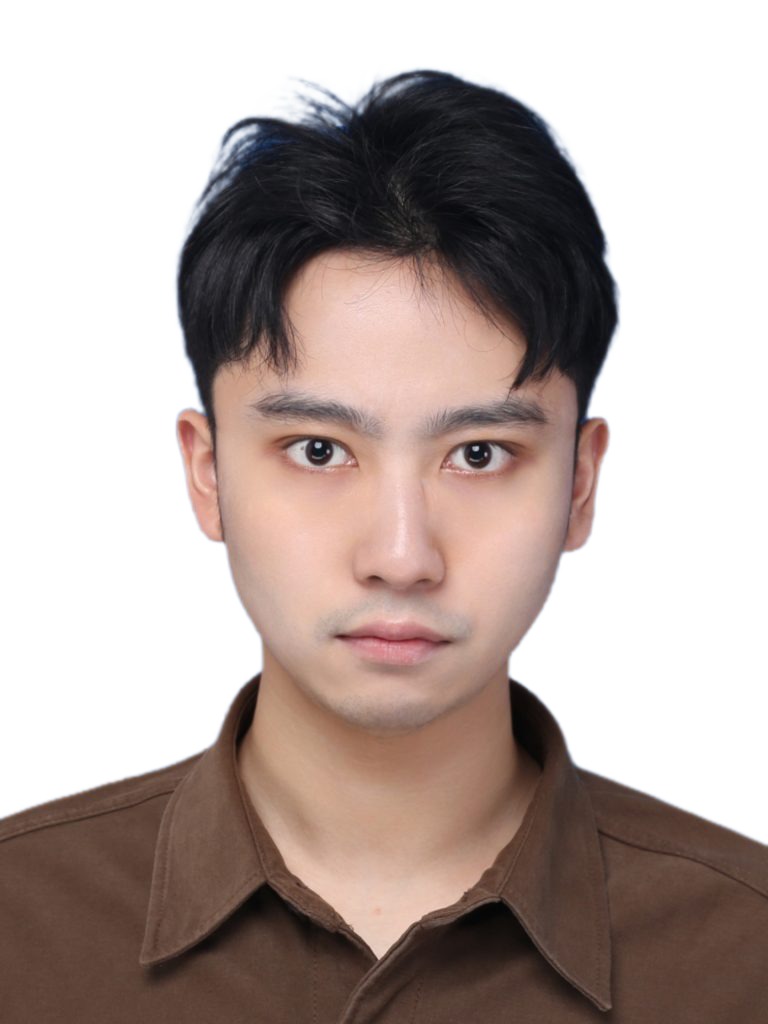}}]{Renwei Yang} received his bachelor’s degree and master’s degree from University of Electronic Science and Technology of China (UESTC), in 2020 and 2023. He is currently pursuing the PhD degree with Department of Computer Science, City University of Hong Kong. His research interests include multimodal large language model, image quality assessment and enhancement, and video coding.
\end{IEEEbiography}

\begin{IEEEbiography}
[{\includegraphics[width=1in,height=1.25in,clip,keepaspectratio]{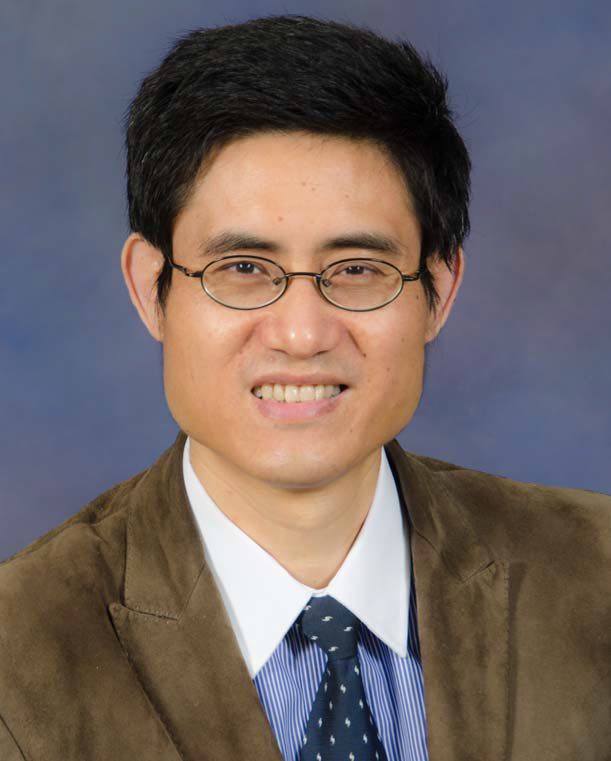}}]
{Dapeng Oliver Wu} (S'98--M'04--SM'06--F'13)  received 
a Ph.D. degree in electrical and computer engineering from Carnegie Mellon University, Pittsburgh,
PA, in 2003.

He is Yeung Kin Man Chair Professor of Network Science, and Chair Professor of Data Engineering at the Department of Computer Science, City University of Hong Kong.  Previously, he was on the faculty of University of Florida, Gainesville, FL, USA and was the director of NSF Center for Big Learning, USA.  His research interests are in the areas of networking, communications, signal processing, computer vision, machine learning, and information and network security. He received University of Florida Term Professorship Award in 2017, University of Florida Research Foundation Professorship Award in 2009, AFOSR Young Investigator Program (YIP) Award in 2009, ONR Young Investigator Program (YIP) Award in 2008, NSF CAREER award in 2007, the IEEE Circuits and Systems for Video Technology (CSVT) Transactions Best Paper Award for Year 2001, and the Best Paper Awards in IEEE GLOBECOM 2011 and International Conference on Quality of Service in Heterogeneous Wired/Wireless Networks (QShine) 2006. 

He has served as Editor in Chief of IEEE Transactions on Network Science and Engineering, Editor-at-Large for IEEE Open Journal of the Communications Society, founding Editor-in-Chief of Journal of Advances in Multimedia, and Associate Editor for IEEE Transactions on Cloud Computing, IEEE Transactions on Communications, IEEE Transactions on Signal and Information Processing over Networks, IEEE Signal Processing Magazine, IEEE Transactions on Circuits and Systems for Video Technology, IEEE Transactions on Wireless Communications and IEEE Transactions on Vehicular Technology.   He has served as Technical Program Committee (TPC) Chair for IEEE INFOCOM 2012, and TPC chair for IEEE International Conference on Communications (ICC 2008), Signal Processing for Communications Symposium, and as a member of executive committee and/or technical program committee of over 100 conferences. 
\end{IEEEbiography}

\vfill

\end{document}